\documentclass[journal]{IEEEtran}

\usepackage[center]{caption}
\usepackage{amsmath}
\usepackage{mathtools}
\usepackage{amsfonts}
\usepackage{citesort}
\usepackage{algorithm}
\usepackage{algpseudocode}
\usepackage{array}
\usepackage{graphicx}
\usepackage{subcaption}
\usepackage{enumitem}
\usepackage{comment}
\newcommand{\fontsmall}{\fontsize{8pt}{9pt}\selectfont}
\usepackage{pgfplots}
\pgfplotsset{compat=1.6}
\usepgfplotslibrary{groupplots}
\usepackage{tabularx}
\usepackage{makecell}
\usepackage{multirow}
\usepackage{colortbl}
\usepackage{rotating}
\algtext*{EndWhile}
\algtext*{EndIf}
\algtext*{EndFor}

\begin{document}
\title{AdaPool: A Diurnal-Adaptive Fleet Management Framework using Model-Free Deep Reinforcement Learning and Change Point Detection}

\author{Marina~Haliem, 
     Vaneet~Aggarwal, 
       and~Bharat~Bhargava
\thanks{The authors are with Purdue University, West Lafayette, IN 47907 USA, email: \{mwadea, vaneet, bbshail\}@purdue.edu.}}


\maketitle

\begin{abstract}
This paper introduces an adaptive model-free deep reinforcement approach that can recognize and adapt to the diurnal patterns in the ride-sharing environment with car-pooling. Deep Reinforcement Learning (RL) suffers from catastrophic forgetting due to being agnostic to the timescale of changes in the distribution of experiences. Although RL algorithms are guaranteed to converge to optimal policies in Markov decision processes (MDPs), this only holds in the presence of static environments. However, this assumption is very restrictive. In many real-world problems like ride-sharing, traffic control, etc., we are dealing with highly dynamic environments, where RL methods yield only sub-optimal decisions. To mitigate this problem in highly dynamic environments, we (1) adopt an online Dirichlet change point detection (ODCP) algorithm to detect the changes in the distribution of experiences, (2) develop a Deep Q Network (DQN) agent that is capable of recognizing diurnal patterns and making informed dispatching decisions according to the changes in the underlying environment. Rather than fixing patterns by time of week, the proposed approach automatically detects that the MDP has changed, and uses the results of the new model. In addition to the adaptation logic in dispatching, this paper also proposes a dynamic, demand aware vehicle-passenger matching and route planning framework that dynamically generates optimal routes for each vehicle based on online demand, vehicle capacities, and locations.  Evaluation on New York City Taxi public dataset shows the effectiveness of our approach in improving the fleet utilization, where less than 50\% of the fleet are utilized to serve the demand of up to 90\% of the requests, while maximizing profits and minimizing idle times.

\end{abstract}

\IEEEpeerreviewmaketitle

\section{Introduction}
In Q-learning, there is a tight coupling between the learning dynamics (probability of choosing an action) and underlying execution policy (the effective rate of updating the Q value associated with that action). This coupling can cause performance degradation in dynamic noisy environments \cite{2-ada}. As the RL agent continues to build on its experiences in order to learn increasingly complex tasks, it should be able to quickly adapt while maintaining its acquired knowledge. However, once the i.i.d. assumption in the inherent Markov Decision Process (MDP) is violated, artificial neural networks have been shown to suffer from \textit{catastrophic forgetting} \cite{8-ada,3-ada} due to erasing knowledge acquired from older data as the model gets trained on the new data. As an example, for an extraterrestrial rover mission, changes in the MDP may be a consequence of regular, predictable events such as intra-day or seasonal temperature variations, or may result from more complex phenomena that are difficult to predict (e.g., terrain changes due to wind) \cite{zimbelman2014introduction}. This paper proposes a novel approach to deal with the model changes in a model-free reinforcement learning setup. 


 We propose an adaptive model-free deep learning framework for ride-sharing with car-pooling that can learn different underlying contexts of the environments. Deep reinforcement learning methodologies are used for this adaptive modeling where transition probabilities are computed through Deep Q Neural Network (DQN).  We utilizes the dispatch of idle vehicles using a Deep Q-learning (DQN) framework as in DeepPool \cite{deep_pool}, and we add the profit term in the reward function so that the output expected discounted rewards (Q-values) associated with each action, becomes a good reflection of the expected earnings gained from performing this action. To the best of our knowledge, our AdaPool framework is the first work that introduces an adaptive model-free approach for distributed matching and dispatching where agents are able to recognize various diurnal patterns, learn their corresponding models, detects the change points and adapts accordingly. Thus, influencing the decision making of ride-sharing platforms. We identify the following as our major contributions: 
\begin{itemize}[leftmargin = *]
	\item We propose an adaptive model-free RL algorithm for handling non-stationary environments, where we adapt Deep Q-learning to learn optimal policies for different environment models. The proposed approach, in training, finds the  set of models that divide the time-varying MDPs based on diurnal patterns, and uses the appropriate model to make decisions. To the best of our knowledge, this is the first work for model-free reinforcement learning that works for training policies accounting for diurnal patterns in any area. 
	\item Using  change point detection, the proposed algorithm switches between the MDP policies, and estimates policy for the new model or improves the policy learnt, if the model had been previously experienced. In this manner, our method avoids \textit{catastrophic forgetting} \cite{8-ada}, and utilizes the Q-values learnt using DQN for making the dispatch decisions.
	\item We propose a dynamic, demand-aware matching and route planning framework, that is scalable up to the maximum capacity per vehicle in the initial assignment phase. In the optimization phase, this algorithm takes into account the near-future demand in order to improve the route-planning by eliminating rides heading towards opposite directions and applying insertion operations to vehicles' current routes. 
	\item  Through experiments using real-word dataset of New York City's taxi trip records \cite{10}  (15 million trips), we simulate the ride-sharing system \footnote{The code for this work is available at: https://github.com/marina-haliem/AdaPool.}. We show that the optimization problem of our novel AdaPool framework is formulated such that it enhances the overall acceptance rate, increases the profit margins of the fleet, minimizes the extra travel distance and the average idle time, when compared to non-adaptive RL approaches. 
\end{itemize}

The rest of this paper is organized as follows: Detailed related work is presented in Section \ref{related}. Section \ref{joint} describes the overall architecture of our adaptive deep RL framework: AdaPool. In Section \ref{darm}, we explain our dynamic, demand-aware approach for matching and route planning. Section \ref{dqn_algo}, explains our adaptive DQN-based approach for dispatching vehicles. Simulation setup as well as experimental results are presented in Section \ref{results}.  Finally, Section \ref{conc} concludes the paper with discussion on future directions.\\
\\
\textit{Abbreviations and Acronyms: }\label{AA}{\small MoD: Mobility-on-Demand. RL: Reinforcement Learning. DQN: Deep Q-Network. ODCP: Online-parametric Dirichlet Change Point. Conv-Net: Convolutional Neural Network. OSRM: Open Source Routing Machine. MDP: Markov Decision Process}.

\section{Related Work}\label{related}
\textbf{Dynamic Ride-sharing:}
In recent years,  dynamic ride-sharing has been  gaining attention.  
Many of the studies on optimal taxi dispatching address the problem as either a variant of the vehicle routing problem (VRP) \cite{jung2016dynamic}; or the bipartite graph matching problem \cite{tafreshian2020trip} where each taxi is matched with the closest passenger in its vicinity.  
The algorithm in \cite{alonso2017demand} generates optimal routes for large scale ridesharing using constrained optimization.  Some approaches consider Tabu search and heuristic algorithms for a robust optimization \cite{li2020ride}. Recently, Reinforcement Learning (RL) has been shown to reach near-optimal solutions that outperform meta-heuristic algorithms \cite{HU2020106244}, especially on large scale problems like ours. \\
In \cite{fleet_oda}, the authors proposed a model free deep RL approach to optimize fleet utilization in ridesharing; however, they don't consider car-pooling.  The authors of \cite{deep_pool} provided the first model-free approach for ridesharing with car-pooling, DeepPool, based on deep RL However, DeepPool does not consider non-stationary environments where the underlying model changes, and thus, does not recognize or adapt to such contextual changes. It primarily focuses on dispatching idle vehicles.  To the best of our knowledge, this is the first work for model-free reinforcement learning that accounts for underlying contextual changes in the environment. 


\textbf{Non-Stationary MDPs:} 
Accounting for time-varying changes in the environment presents a major challenge in decision making. The authors of \cite{nilim2005robust} consider MDPs with arbitrarily changing state-transition probabilities but fixed reward functions where it is assumed that the uncertainty in the transition probabilities is state-wise independent. The authors of \cite{even2005experts,dick2014online} consider the problem of MDPs with fixed state-transition probabilities and changing reward functions. The authors of \cite{ortner2020variational} consider the case where both the reward functions and the state-transition probabilities may vary
(gradually or abruptly) over time. In literature, most approaches that address catastrophic forgetting focus on sequential learning of distinct tasks, where they rely on the awareness of task boundaries \cite{4-ada}, \cite{5-ada}, \cite{6-ada}. This is not practical because in many situations the data distribution evolve gradually over time during training, and thus can not be discretized into separate tasks. Other approaches for time-varying MDPs have been studied, for details see \cite{liu2018solution,ornik2019learning}. However, these approaches are not applicable for dividing training and testing separately and cannot be used for diurnal patterns where even though the model has changed, we might have seen such a model in the past. In the presence of diurnal patterns, we consider that the model changes to be close to one of the models we have seen in the past. However, detecting that there is a change of MDP and then using (and fine-tuning) a model in the past is important with diurnal patterns, and is the focus of this work. We discretize the number of possible models to a constant and the proposed framework learns the efficient split of the diurnal changing MDP into a set of models with efficient change point. This is akin to clustering MDPs into certain categories and using the relevant MDP based on the change point detection. We note that this division is done without estimating the transition probabilities of the MDP, and is thus a model-free approach. 

\textbf{Change Detection:} It is a non-trivial problem to detect unexpected changes and adapt quickly to them, without giving any prior knowledge to the system about such changes in advance (e.g., peak, non-peak demand time periods, city holidays, daylight savings, etc.). Also, whether conditions can introduce significant unexpected changes to the system. Snow can shift the schools/colleges and have delayed start. {In \cite{wang2020multiscale}, the authors presented a multi-scale drift detection framework that localizes abrupt drift points on two different scales. The authors of \cite{zhang2015collaborated} proposed a collaborated online change-point detection method to leverage engagement metrics for detecting sequential changes in spare time series.  In \cite{chen2018parallel},  the authors  utilize a generative adversarial network (GAN) to learn from virtual emergencies generated in artificial traffic scenes in addition to the real ones. However, this aims to detect only emergency situations rather than learning the underlying pattern of change which is the goal for this paper.
	
	In this paper, we approach this problem by introducing AdaPool - An Adaptive deep reinforcement learning approach that can learn several models associated with different diurnal patterns instead of learning only one static model in ride-sharing environment with car-pooling. Our approach performs efficient change point detection, learns the new model or improves the existing model according to the samples it observes, and adapts accordingly. We note that the proposed approach can be used for exploitation after learning for diurnal patterns where no new model is learnt and the different learnt models are exploited with change point detection, unlike that in \cite{nilim2005robust,even2005experts,dick2014online,ortner2020variational,liu2018solution,ornik2019learning}. Finally, the time of changes are not pre-programmed and thus the approach is adaptable to the dynamic changes like city holidays. To the best of our knowledge, this is the first work for model-free reinforcement learning that works for training policies accounting for diurnal patterns in any area. 
	
	\textbf{Route Planning:} Another problem addressed in this paper is \textit{route planning}, which has been proven to be an NP-hard problem \cite{6,DARP}. Route planning - given a set of vehicles $V$, and requests $R$ - designs a route that consists of a sequence of pickup and drop-off locations of requests assigned to each vehicle.  In \cite{jiang2019path}, the authors utilize DQN with experience replay for path planning. However, their goal is only for robots to avoid collisions with static objects. } In ride-sharing environments, vehicles and requests arrive dynamically and are arranged such that they meet different objectives (e.g., maximizing the the number of served requests \cite{14}). 
Utilizing an operation called \textit{insertion} to solve such a highly dynamic problem has been proven, in literature, to be both effective and efficient \cite{12,14,32,38,40-37}. The \textit{insertion} operation aims to accommodate the newly arriving request by inserting its pickup (i.e., origin) and its drop-off (i.e., destination) locations into a vehicle’s route.  Recent works (e.g., \cite{darm}) have also integrated pricing with route-planning in order not to plan opposite destination passengers together, while efficient pricing with non-stationary setup is left for future work.
\\
{The authors of \cite{9003511} propose an arc routing approach to solve the garbage collection problem based on minimization of service cost, while the authors in \cite{8388738} present an iterative heuristic and a fuzzy logic method to solve the food logistics problem. In \cite{8651901}, the authors present a memetic algorithm with competition (MAC) to solve the capacitated green vehicle routing problem. }
However, most of the approximation algorithms that provide solutions to the route planning problem in ridesharing are limited to only two requests per vehicle (e.g., \cite{6,DARP}.) To mitigate this issue, we adopt an insertion-based route-planning framework similar to \cite{darm}. However, in \cite{darm}, the initial assignment phase is restricted by the vehicle's capacity constraint, which could lead to lower acceptance rate due to not considering other potential rides. We improve this approach by allocating up to 50 potential assignments to each vehicle without restricting to its capacity. Then, we optimize by performing greedy-insertion operations to decide on the $k^{\prime}$ rides that satisfy the capacity constraint while achieving the route corresponding to the minimum insertion cost possible among all potential requests in vicinity. 

\vspace{-0.1in}
\section{AdaPool: Adaptive Matching and Dispatching Framework using Deep RL}\label{joint}
\begin{figure}
\captionsetup{justification=centering, font=small, format=hang}
\centering
\includegraphics[trim = 50 200 50 150, width=.35\textwidth]{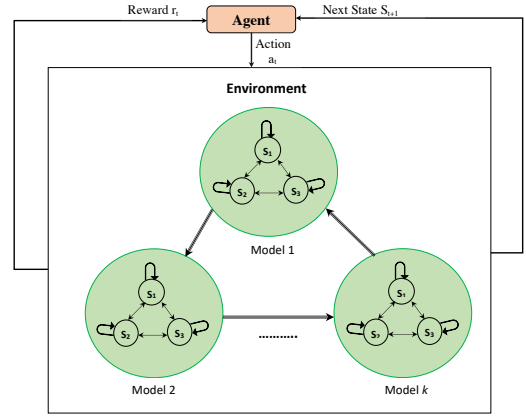}
\vspace{-1em}
\caption{\small Time Varying MDPs in Non-Stationary Environment} \label{dynamic}
\vspace{-1em}
\end{figure}

In this section, we provide details about each component in our model architecture,  as well as explanations of model parameters and notations. We propose a novel adaptive framework for matching and dispatching in ride-sharing environments with car-pooling using DQN, where initial matchings (that are decided in a greedy fashion) are then optimized in a distributed manner (per vehicle) in order to meet the vehicle's capacity constraints as well as minimize customers' extra waiting time and driver's additional travel distance. We consider the scenario where the environment changes between models $1, 2, \cdots, k$ in a cyclic manner, dynamically as shown in Fig. \ref{dynamic}. With the cyclic nature, the environment changes from model $n$ to model $1$. Such cyclic repetition allows for learning the $k$ models and use them in the run-time, where on detecting the change-point, next model can be used. The implication of the non-stationary environment is this: when the agent exercises a control $a_{t}$ at time $t$, the next state $s_{t+1}$ as well as the reward $r_{t}$ are functions of the \textit{active} environment model dynamics. In our approach, we assume that there are $k$ environment models $M_{1}, M_{2}, \cdots, M_{k}$, through which the system cycles. However, neither the context information 
(or model parameters) of each model nor the change points $T_{1}, T_{2}, \cdots$  (when these model changes occur), are known to the RL agent. In this case, the agent can collect experience tuples while simultaneously following a model-free learning algorithm to learn an approximately optimal policy. Instead of assuming any specific structure, our model-free approach learns the Q-values dynamically using convolutional neural networks. Our method works in tandem with a change point detection algorithm, to get information about the changes in the environment. Then, it updates Q-values of the relevant model whenever a change is detected and does not attempt to estimate the transition and reward functions for the new model. Additionally, if the method finds that samples are obtained from a previously observed model, it updates the Q values corresponding to that model. Thus, in this manner, the information which was learnt and stored earlier (in the form of Q-values) is not lost.

Moreover,  each vehicle learns the best future dispatch action to take at time step $t$, taking into consideration the locations of all other nearby vehicles, but without anticipating their future decisions. Vehicles' dispatch decisions are made in parallel, since it is unlikely for two drivers to take actions at the same exact time since drivers know the location updates of other vehicles in real time (e.g., GPS). 
Therefore, our algorithm learns the optimal policy for each agent independently as opposed to centralized-based approaches such as in \cite{fleet_oda}. 

\vspace{-0.1in}
\subsection{Model Architecture}
\begin{figure}
\captionsetup{justification=centering, font=small, format=hang}
\centering
\includegraphics[trim = 30 50 50 50, width=.45\textwidth]{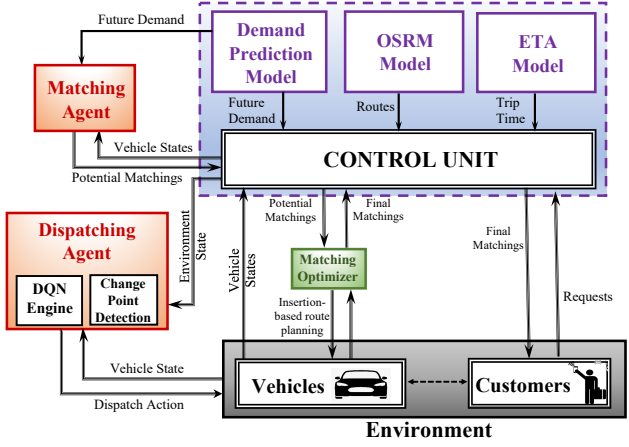}
\vspace{-1em}
\caption{\small Overall architecture of AdaPool framework} \label{arch}
\vspace{-1em}
\end{figure}

Figure \ref{arch} shows the basic components of our joint framework and the interactions between them. We assume that the control unit is responsible for: (1) making the initial matching decisions, based on the proximity of vehicles to ride requests, (2) maintaining the states including current locations, current capacity, destinations, etc., for all vehicles. These states are updated in every time step based on the dispatching and matching decisions. (3) Control unit also has some internal components that help manage the ride-sharing environment such as: (a) the estimated time of arrival (ETA) model used to calculate and continuously update the estimated arrival time. (b) The Open Source Routing Machine (OSRM) model used to generate the vehicle's optimal trajectory, using \cite{osm}, to reach a destination, and (c) the (Demand Prediction) model used to calculate the future anticipated demand in all zones. We adopt these three models from \cite{fleet_oda,dprs} since they play a crucial role in enabling this framework; their details are provided in Appendix \ref{models}.
First, the ride requests are input to the system along with the heat map for supply and demand (which involves demand prediction in the near future).  {Then, based on the predicted demand, vehicles adopt a dispatching policy using DQN, where they get dispatched to zones with anticipated high demand. This step not only takes place when vehicles first enter the market (lines 3-5 in Algorithm \ref{joint_algo}), but also when they experience large idle durations (at the end of every time step, this gets checked for in lines 13-14, Algorithm \ref{joint_algo}).} The control unit performs the initial vehicle-passenger(s) matching where each vehicle gets assigned all potential (one or more) requests in its vicinity.   {Then, 
each vehicle executes its matching optimizer module that performs an insertion-based route planning.  In this step, vehicles reach their final $n$ requests by dealing with their initial matchings list in the order of their proximity, performing an insertion operation to their current route plan (as long as this insertion satisfies the capacity, extra waiting time, and additional travel distance constraints to guarantee that serving this request would yield a profit). }

Using the ``Change Point Detection" module, vehicles are able to learn and switch between models $M_{1}, M_{2}, \cdots, M_{k}$ dynamically whenever a change is detected. Finally, a vehicle communicates with the control unit, as needed, to request new information of the environment (prior to making decisions) or update its own status (after any decision). 
Algorithm \ref{joint_algo} presents the overall flow of our framework.

\begin{algorithm}
	\caption{ \small AdaPool Framework} \label{joint_algo}
	\begin{minipage}{\linewidth}
		\fontsmall
		\begin{algorithmic}[1]
			\State \textbf{Initialize} vehicles' states $X_{0}$ at $t_{0}$.
			\For {$t \in T$}
				\State \textbf{Fetch} all vehicles that entered the market in time slot t, $V_{new}$.
				\State \textbf{Initialize} Vehicles' routes $S_{V_{j}} \gets $ empty for each $V_{j} \in V_{new}$
				\State \textbf{Dispatch} $V_{new}$ to zones with anticipated high demand (Algo. \ref{dqn_brief}).
				\State \textbf{Fetch} all ride requests at time slot t, $D_{t}$.
				\State \textbf{Fetch} all available vehicles at time slot t, $V_{t}$.	
				\For {each vehicle $V_{j} \in V_{t} \ldots$}
					\State \textbf{Obtain} initial matching $A_{j}$ using Algorithm \ref{assign}.
					\State \textbf{Perform} route planning $S_{V_{j}} \gets$ \textsc{Greedy\_Insertion}{($A_{j}, S_{V_{j}}$)}
					\State \textbf{Retrieve} next stop from $S_{V_{j}}$.
					\State \textbf{Head} to next stop (whether a pickup or a dropoff).
				\EndFor
				\State \textbf{Fetch} all idle vehicles with $\text{Idle\_duration} > 10$ minutes, $V_{idle}$.
				\State \textbf{Dispatch} $V_{idle}$ to zones with anticipated high demand (Algo. \ref{dqn_brief}).
				\State \textbf{Update} the state vector $s_{t}$.
			\EndFor
			\Procedure{Greedy\_Insertion}{$A_{j}, S_{V_{j}}$}
				\State \textbf{Initialize} $V^{j}_{\text{capacity}} = V^{j}_{C}$
				\While{$V^{j}_{\text{capacity}} < C^{V_{j}}_{max}$}
					\For {each ride request $r_{i} \in A_{j} \ldots$}
						\If{$(V^{j}_{\text{capacity}} + |r_{i}|) \leq C^{V_{j}}_{max}$}
							\State \textbf{Obtain} $(S^{\prime}_{V_{j}}, \text{cost(}V_{j}, S^{\prime}_{V_{j}})) \gets$ 			 								\textsc{route\_planning} ($V_{j}, S_{V_{j}}, r_{i}$) using Algo. \ref{insert}.
						\EndIf
					\EndFor
					\State min\_cost $\gets min_{(r_{i} \in A_{j})}(\text{cost(} V_{j}, S^{\prime}_{V_{j}}[r_{i}]))$).
					\State $r^{*} \gets argmin_{(r_{i} \in A_{j})}(\text{cost(} V_{j}, S^{\prime}_{V_{j}}[r_{i}]))$).
					\State \textbf{Update} trip time $T_{i}$ based on $S^{\prime}_{V_{j}}[r^{*}]$ using ETA model.
					\State \textbf{Update} $S_{V_{j}} \gets S^{\prime}_{V_{j}}[r^{*}]$
					\State \textbf{Increment} $V^{j}_{\text{capacity}} \gets V^{j}_{\text{capacity}} + |r^{*}|$
					\State \textbf{Remove} $r^{*}$ from $A_{j}$
					\If{$A_{j}$ is \textit{empty}}
						\State \textbf{\textit{break}}
					\EndIf
				\EndWhile
				\State \textbf{Update} the state vector $s_{t}$.
			\EndProcedure
		\end{algorithmic} 
	\end{minipage}
\end{algorithm}
\vspace{-0.1in}
\subsection{Model Parameters and Notations}\label{MP}
We built a ride-sharing simulator to train and evaluate our framework. We simulate New York City as our area of operation, where the map is divided into multiple non-overlapping regions, a grid with each 1 square mile being taken as a zone. This allows us to discretize the area of operation and thus makes the action space\textemdash where to dispatch the vehicles\textemdash tractable. We use $m \in \{1,2,3, ..., M\}$ to denote the city's zones, and $n$ to denote the number of vehicles. A vehicle is marked as \textit{available} if there is remaining seating capacity. Vehicles that are completely full or are not considering taking passengers are marked \textit{unavailable}. Available vehicles in zone $i$ at time slot $t$ is denoted $v_{t,i}$. Only available vehicles are eligible to be dispatched. We optimize our algorithm over $T$ time steps, each of duration $\Delta t$. The vehicles make decisions on where on the map to head-to to serve the demand at each time step $\tau = t_{0},t_{0}+\Delta t,t_{0}+2(\Delta t),\ldots,t_{0}+T(\Delta t)$ where $t_{0}$ is the start time.  {Below, we present the model parameters and notations}:
\begin{itemize}[leftmargin=*]
	\item \textit{Demand:} We denote the number of requests for zone $m$ at time $t$ as $d_{t,m}$. The future pick-up request demand in each zone is predicted through a historical distribution of trips across the zones \cite{plan_article}, and is denoted by $D_{t:T} = (\boldsymbol{\overline{d}_{t}},\ldots,\boldsymbol{\overline{d}_{t+T}})$ from time $t$ to $t + T$.
	
	\item \textit{Supply:} At each time slot $t$, the supply of vehicles for each zone is projected to future time $\tilde{t}$. $d_{t,\tilde{t}, m}$ is the number of vehicles that are currently unavailable at time $t$ but will become available at time $\tilde{t}$ as they will drop-off customer(s) at region $m$. This information can be ascertained using the ETA \cite{deep_pool, dprs} prediction for all vehicles. Consequently, for a set of dispatch actions at time $t$, we can predict the number of vehicles in each zone for $T$ time slots ahead, from time $t$ to time $t+T$, denoted by $V_{t:t+T}$ which serves as our predicted supply in each zone for $T$ time slots ahead. 
	
	\item \label{state} \textit{Vehicle Status:} 
	 We use $X_{t} = \{x_{t,1}, x_{t,2}, \; ... \; , x_{t,N}\}$ to denote the $N$ vehicles' status at time $t$. $x_{t,n}$ tracks vehicle $n$'s state variables at time step $t$ including, (1) current location/zone $V_{loc}$,  (2) current capacity $V_{C}$, (3) type $V_{T}$,  (4) pickup time of each at passenger,  (5) the destination  of each passenger, and (6) the earnings till time $t$. A vehicle is considered available, if and only if $V_{C} < C_{max}^{V}$. 
\end{itemize}

These variables change in real time according to the environment variations and demand/supply dynamics. However, our framework keeps track of all these rapid changes and seeks to make the demand, $d_{t}$, $\forall t$ and supply $v_{t}$, $\forall t$ close enough (i.e., mismatch between them is zero). Combining all this data, we define the state space, action space and reward function for our DQN agents:
\begin{enumerate}[leftmargin=*]
	\item \textit{State Space:} we have defined a three tuple that captures the environment updates at time $t$ to represent our state space as $s_{t} = (X_{t}, V_{t:t+T}, D_{t:t+T})$. When a set of new ride requests arrive at the system, we can retrieve from the environment all the state elements, combined in one vector $s_{t}$. Also, when a passenger's request becomes assigned, we append the customer's expected pickup time, source, destination and ride fare to $s_{t}$ as well. 
	
	\item \label{actions} \textit{Action Space:} $a_{t}^{n}$ denotes the action taken by vehicle $n$ at time step $t$. In our simulator, the vehicle can move (vertically or horizontally) at most 7 cells, and hence the action space is limited to these cells. A vehicle can move to any of the 14 vertical (7 up and 7 down) and 14 horizontal (7 left and 7 right). This results in a 15x15 action space $a_{t, n}$ for each vehicle as a vehicle can move to any of these cells or it can remain in its own cell. After the vehicles decides on which cell to go to using DQN (Section \ref{ada_dqn}), it uses the shortest optimal route to reach its next stop.
	
	\item \textit{Reward:} Having explained all of the above factors, at every time step $t$, the DQN agent obtains a representation for the environment, $s_{t}$, and a reward $r_{t}$. Based on this information, the agent takes an action that directs the vehicle (that is either idle or recently entered the market) to different dispatch zone where the expected discounted future reward is maximized, i.e., $
	\sum^{\infty}_{j = t} \eta^{j-t} r_{j}(a_{t}, s_{t})$, where $\eta < 1$ is a time discount factor. In our algorithm, we define the reward $r_{k}$ as a weighted sum of different performance components that reflect the objectives of our DQN agent (explained in Section \ref{dispatch}). The reward will be learnt from the environment for individual vehicles and then leveraged to optimize their decisions. 
\end{enumerate}

We note that this discount factor is a key that makes the change of model slow to learn. Thus, we use change point detection to learn the change and use the appropriate model. A detailed table of notations is provided in Appendix \ref{apd:notations}.

\vspace{-0.1in}
\section{Dynamic Demand-Aware Matching and Route-Planning Framework}\label{darm}
This section provides details of our dynamic, demand-aware approach to solve the NP-Hard matching and route planning problems  in ride-sharing environments. Our framework goes through two phases as explained in this section.

\textbf{NP-Hardness:}  The ride-sharing assignment problem is proven to be NP-hard in \cite{6} as it is a reduction from the 3-dimensional perfect matching problem (3DM). The authors of \cite{6} provided  an approximation algorithm that is 2.5 times the optimal cost for the case where at most two requests can share the same vehicle at a time. However, our approach is not limited to at most two requests per vehicle. The proposed approach is a different from  that in \cite{darm} where pricing was used to dis-incentivize matching passengers going in opposite directions, while we will use  a greedy approach in adding the customers to the route. 



\vspace{-0.15in}
\subsection{Initial Vehicle-Passenger(s) Assignment Phase} \label{phase1}

Note that the control unit for decision making knows  the future demand $D_{t:t+T}$ at each zone, the vehicles' status vectors $X_{t}$ including their current locations and destinations as well as the origin $o_{i}$ and destination $d_{i}$ locations for each request $r_{i}$. Each vehicle is assigned to up to $50$  requests $r_{i}$ in its vicinity (to reduce the computational power needed), that could potentially get served by it. At the end of this phase, each vehicle $V_{j}$ has a list of initial matchings $A_{j} = [r_{1}, r_{2}, ..., r_{k}]$, where $k \leq 50$ (Pseudo-code for this phase is given in Appendix \ref{apd:match}, Algorithm \ref{assign}). 

\vspace{-0.15in}
\subsection{Optimization Phase: Greedy Insertion Cost}\label{phase2}

In this phase, we follow the idea of searching each route and locally optimally inserting new vertex (or vertices) into a route. In our problem, there are two vertices (i.e., origin $o_{i}$ and destination $d_{i}$) to be inserted for each request $r_{i}$. We define the insertion operation as: given a vehicle $V_{j}$ with the current route $S_{V_{j}}$,  and a new request $r_{i}$, the insertion operation aims to find a new feasible route $S^{\prime}_{V_{j}}$ by inserting $o_{i}$ and $d_{i}$ into $S_{V_{j}}$ with the minimum increased cost, that is the minimum extra travel distance, while maintaining the order of vertices in $S_{V_{j}}$ unchanged in $S^{\prime}_{V_{j}}$. 

Specifically, for a new request $r_{i}$, the basic insertion algorithm checks every possible position to insert the origin and destination locations and return the new route such that the incremental cost is minimized. So, the vehicle reaches its final matchings (denoted $M_{j}$) list by greedily picking the top $k^{\prime}$ requests (where $k^{\prime} \subset k$) with the minimal insertion cost while $k^{\prime} \leq C^{V_{j}}_{max}$ to satisfy its capacity constraint (see lines 16-24 in Algorithm \ref{joint_algo}). 
Assume the passenger count per request is $\mid r_{i} \mid $, and the vehicle $V_{j}$ arrives at location $z$. Then, to check the capacity constraint in $O(1)$ time, we define vehicle $V_{j}$'s current capacity $V^{j}_{C}[z]$  which refers to the total capacity of the requests that are still on the route of $V_{j}$ when it arrives at that location $z$ as follows:
$
\fontsmall
V^{j}_{C}[z]  =
\begin{cases}
V^{j}_{C}[z-1] + \mid r_{i} \mid  & \text{if $z == o_{i}$} \\
V^{j}_{C}[z-1] - \mid r_{i} \mid  &  \text{if $z == d_{i}$} 
\end{cases}
$.

To present our cost function, we first define our distance metric, where given a graph $G$, we use our OSRM engine to pre-calculate all possible routes over our simulated city. Then, we derive the distances of the trajectories (i.e., paths) from location $a$ to location $b$ to define our graph weights. Thus, we obtain a weighted graph $G$ with realistic distance measures serving as its weights. We extend the weight notation to paths as follows:
 $w(a_{1}, a_{2}, ..., a_{n}) = \sum_{i=1}^{n-1} w(a_{i}, a_{i+1})$.  Thus, we define the cost associated with each new potential route/path $S^{\prime}_{V_{j}} = [r_{i}, r_{i+1}, ..., r_{k}]$ to be the $\text{cost}(V_{j}, S^{\prime}_{V_{j}}) = w(r_{i}, r_{i+1}, ... r_{k})$ resulting from this specific ordering of vertices (origin and destination locations of the $k$ requests assigned to vehicle $V_{j}$). 
 The full insertion-based algorithm is presented in Algorithm \ref{insert}. 

After vehicle $V_{j}$ picks the route $S^{\prime}_{V_{j}}$ that has the minimum cost of inserting $r_{y}$ into its current route, it follows the same procedure for every $r_{i} \in A_{j}$ as shown in \textit{Greedy\_Insertion} procedure in Algorithm \ref{joint_algo}. Repeatedly picking $r^{*}$ that is the \textit{argmin} of the minimum insertion cost among all potential requests in $A_{j}$, each vehicle ends up with $k^{\prime}$ final matchings that guarantees the optimal routing for the vehicle while serving all $k^{\prime}$ requests and satisfying its capacity constraint. 

Finally, this phase works in a distributed fashion where each vehicle picks the top $k^{\prime}$ requests that minimizes its travel cost, following Algorithm \ref{insert}, and satisfying its capacity constraint by following the \textit{Greedy\_Insertion} procedure in Algorithm \ref{joint_algo}. 

\textbf{Complexity Analysis:} We note that the routes  pre-calculation step done using our OSRM engine (with their associated costs), provides us with fast routing and constant-time computation $O(1)$, thus reducing the complexity of our algorithm from $O(n^{3})$ to $O(n^{2})$.  In addition,  we adopt the approach proposed in \cite{darm} for checking the route feasibility in $O(1)$ time to further reduce the computation needed (details provided in Appendix \ref{apd:match}).

\begin{algorithm}[ht]
	\caption{Insertion-based Route Planning} \label{insert}
	\begin{minipage}{.95\linewidth}
		\fontsmall
		\begin{algorithmic}[1]
		\State \textbf{Input: } Vehicle $V_{j}$, its current route $S_{V_{j}}$, a request $r_{i} = (o_{i}, d_{i})$ and weighted graph G with pre-calculated trajectories using OSRM model.
		\State \textbf{Output: } Route $S^{\prime}_{V_{j}}$ after insertion, with minimum cost($V_{j}, S^{\prime}_{V_{j}}$). 
		\Procedure{route\_planning}{$V_{j}, S_{V_{j}}, r_{i}$}
		\If { $S_{V_{j}}$ is \textit{empty}}
			\State $S^{\prime}_{V_{j}}  \gets [loc(V_{j}), o_{i}, d_{i}]$.
			\State cost($V_{j}, S^{\prime}_{V_{j}}$) = $w(S^{\prime}_{V_{j}}) $.
			\State \textbf{Return} $S^{\prime}_{V_{j}}$, cost($V_{j}, S^{\prime}_{V_{j}}$)
		\EndIf
		\State \textbf{Initialize} $S^{\prime \prime}_{V_{j}} = S_{V_{j}}$, $Pos[o_{i}] = \text{NULL}$, $\text{cost}_{min} = +\infty$.
		\For {each $x$ in $1$ to $|S_{V_{j}}|$}
			\State $S^{x}_{V_{j}}\; :=$ \textbf{Insert} $o_{i}$ at $x-th$ in $S_{V_{j}}$.
			\State \textbf{Calculate} cost($V_{j}, S^{x}_{V_{j}}$) = $w(S^{x}_{V_{j}})$.
			\If {cost($V_{j}, S^{x}_{V_{j}}) < \text{cost}_{min}$}
				\State $\text{cost}_{min} \gets$ cost($V_{j}, S^{x}_{V_{j}}$).
				\State $Pos[o_{i}] \gets x$, $S^{\prime \prime}_{V_{j}} \gets S^{x}_{V_{j}}$.
			\EndIf
		\EndFor
		\State $S^{\prime}_{V_{j}} = S^{\prime \prime}_{V_{j}}$, $\text{cost}_{min} = +\infty$.
		\For {each $y$ in $Pos[o_{i}] + 1$ to $|S^{\prime \prime}_{V_{j}}|$}
			\State $S^{y}_{V_{j}}\; :=$ \textbf{Insert} $d_{i}$ at $y-th$ in $S^{\prime \prime}_{V_{j}}$.
			\State \textbf{Calculate} cost($V_{j}, S^{y}_{V_{j}}$) = $w(S^{y}_{V_{j}})$.
			\If {cost($V_{j}, S^{y}_{V_{j}}) < \text{cost}_{min}$}
				\State $\text{cost}_{min} \gets$ cost($V_{j}, S^{y}_{V_{j}}$).
				\State $S^{\prime}_{V_{j}} \gets S^{y}_{V_{j}}$, cost($V_{j}, S^{\prime}_{V_{j}}) \gets \text{cost}_{min} $.
			\EndIf
		\EndFor
		\State \textbf{Return} $S^{\prime}_{V_{j}}$, cost($V_{j}, S^{\prime}_{V_{j}}$)
		\EndProcedure
		\end{algorithmic} 
	\end{minipage}
\end{algorithm}

\vspace{-1em}
\section{Adaptive DQN Dispatching Approach} \label{dqn_algo}
In this section, we present our distributed adaptive approach for dispatching vehicles. This framework aims at re-balancing vehicles over the city to better serve the demand while accounting for the different diurnal patterns during the day. Utilizing DQNs along with a change point detection algorithms, individual agents (i.e., vehicles) are able to learn different underlying models of the environment that correspond to the different demand patterns and switch between them according to the observed state of the environment. We utilize a reinforcement learning framework, with which we can learn the probabilistic dependence between vehicle actions and the reward function thereby optimizing our objective function. 
We utilize this framework in order to re-balance vehicles over the city to better serve the demand. The fleet of autonomous vehicles were trained in a virtual spatio-temporal environment that simulates urban traffic and routing. In our simulator, we used the road network of the New York City Metropolitan area along with a realistic simulation of taxi pick-ups. This simulator hosts each deep reinforcement learning agent which acts as a delivery vehicle in the New York City area that is looking to maximize its reward defined by Eq. \eqref{individual}. The learning begins by obtaining experience tuples $E_{t}$ according to the dynamics and reward function of current active model $M_{\theta_{c}}$. The state and reward obtained are stored as experience tuples, since model information is not known. 

\vspace{-1em}
\subsection{Distributed Adaptive DQN } \label{ada_dqn}
At every time step $t$, our  adaptive DQN performs the change point detection algorithm described in Section \ref{cpd}. If it receives $T^{*}$ signalling that a change has been detected, it increments the counter $c$ and starts switching from its current model and updates (and takes action based on) a new model, it does not attempt to estimate the transition and reward functions for the new model. Instead, it starts to update the dynamics of this new model, where the  Q values are updated. The full algorithm for this approach is in Algorithm \ref{dqn_brief}, where as assume the knowledge of a pattern of change as in line 1, but without the knowledge of the context information of each model.

At every time step $t$, the DQN agent obtains a representation for the environment, $s_{t, n}$, and calculates a reward $r_{t}$ associated with each dispatch-to location in the action space $a_{t, n}$ according to the dynamics and reward function of current active model $M_{\theta_{c}}$, and updates Q-values of the relevant model. Based on the rewards associated with each cell of the vehicle's action space explained in Section \ref{MP}, the agent takes an action that directs the vehicle to different dispatch zone where the expected discounted future reward is maximized. In our algorithm, we define the reward $r_{k}$ as a weighted sum of different performance components that reflect the objectives of our DQN agent (explained in Section \ref{dispatch}). The architecture of our DQN is described in Appendix \ref{DQN_Arch}.


\begin{algorithm}[!htb]
	\caption{Distributed Dispatching using Adaptive DQN} \label {dqn_brief}
		\fontsmall
		\begin{algorithmic}[1]
		\State \textbf{Input 1: } Model Change Pattern, $M_{\theta_{1}} \rightarrow M_{\theta_{2}}, M_{\theta_{2}}  \rightarrow M_{\theta_{3}},   \dots, M_{\theta_{k-1}} \rightarrow M_{\theta_{k}}$, where $M_{\theta_{i}} \in \{M_{1}, M_{2}, \dots, M_{k} \}$, and $\theta \in \Theta =  \{1,2, \dots, k\}$
		\State \textbf{Input 2: } $X_{t}, V_{t:t+T}, D_{t:t+T}$.
		\State \textbf{Output: } Dispatch Decisions
		\State \textbf{Fix} learning rate $\sigma$
		\State \textbf{Initialize} context number, c = 1, Q values $Q(m, s, a) = 0$, $\forall \; m \in 1, \dots, k$ $\forall(s,a) \in S$ x $A$.
		\State \textbf{Construct} a state vector $s_{t, n} = (X_{t}, V_{t:t+T}, D_{t:t+T})$.
		\State \textbf{Get} the best dispatch action $a_{t, n} = argmax[Q(s_{t, n}, a; \theta_{c} )]$ for all vehicles $V_{n}$ using the Q-network of model $M_{\theta_{c}}$.
		\State \textbf{Get} the destination zone $Z_{t,j}$ for each vehicle $j \in V_{n}$ based on action $a_{t, j} \in a_{t, n}$
		\State \textbf{Get} reward $r_{t, n}$ using Eq. \ref{individual} associated with model $M_{\theta_{c}}$.
		\State \textbf{Update} Q value associated with model $M_{\theta_{c}}$ (explained in Appendix \ref{learn}).
		\State \textbf{Obtain} next state $s_{t+1, n}$ according to the environment dynamics.
		\State $e_{t} \gets (s_{t,n}, r_{t,n}, s_{t+1, n})$
		\State \textbf{Update} dispatch decisions by adding $(j, Z_{t,j})$
		\State $\tau \gets$ \textsc{DCP} $(e_{T} = {e_{1}, ... e_{t}})$, where T includes all $t \geq \tau^{*}$ at which model $M_{\theta_{c}}$ was active.
		\If{$\tau$ is not Null}
			\State \textbf{Increment} $c = \textit{mod (}c+1, k)$.
			\If{$c == 0$}
			\State $c = k$
			\EndIf
			\State $\tau^{*} \gets \tau$
		\EndIf
		\State \textbf{Return} $(n, Z_{t,n})$ 
		\end{algorithmic} 
\end{algorithm}

The reward will be learnt from the environment for individual vehicles and then leveraged by the agent/optimizer to optimize its decisions. Through learning the probabilistic dependence between the action and the reward function that is explained further in Appendix \ref{learn}, we learn the Q-values according to the dynamics and reward function of current active model $M_{\theta_{c}}$ associated with the probabilities $P(r_{t}\mid a_{t},s_{t})$ over time by feeding the current states of the system. The Q-values are then used to decide on the best dispatching action to take for each individual vehicle. 
Looking at Figure \ref{dynamic}, the DQN agent starts by learning model 1, where c = 1 in line 5 of Algorithm \ref{dqn_brief}. At each time step $t$, it receives the 3-tuple representation of the environment, calculates the reward (Q-values) according to the dynamics of that active model, and makes dispatch decisions accordingly (see lines 7 - 11) by picking the actions that yields the maximum expected discounted reward (Q-value). Besides, the agent stores experience tuples $E_{t}$, at each time step $t$, that consists of current state $s_{t}$, reward $r_{t}$, and next state $s_{t+1}$ as shown in line 12. Further, after the learning step, the agent checks for change points using the ODCP algorithm (line 14) explained in Section \ref{cpd}. Once it detects a change, that is when the ODCP algorithm (Algorithm \ref{sim}) returns $T^{*}$, it switches to next model $c+1$ and continues its learning using the dynamics of the new active model (lines 14 - 19). Note that if the samples observed come from a model (i.e., policy) that has been learnt before, the DQN agent updates the Q-values of that previously seen model and continues learning building on its previous experience that is associated to that model. Also, after learning (where no new model is learnt), the different learnt models can be exploited along with change point detection for recognizing diurnal patterns.

\vspace{-1.2em}
\subsection{Online Dirichlet Change Point Detection} \label{cpd}
To detect points of change, our DQN agents analyze data from their experience memory. The samples can be analyzed for context changes in batch mode or online mode. If a change gets detected, then the counter $c$ is incremented, signalling that the agent believes that context has changed. We adapt the online parametric Dirichlet change-point (ODCP) detection algorithm proposed in \cite{7-ada} for data consisting of experience tuples.  Multiple change-points are detected by performing a sequence of single change-point detections. Although ODCP requires the multivariate data to be i.i.d. samples from a distribution, the justification in \cite{9-ada} explains the utilization of ODCP in the Markovian setting, where the data obtained does not consist of independent samples. The full algorithm for the Dirichlet change point detection algorithm is shown in Algorithm \ref{sim}. In this algorithm, the maximum likelihood estimation of Dirichlet distribution parameters is calculated for the cumulative data (stored through experience tuples) using Eq. \ref{drichlet} below:
\begin{equation} \label{drichlet}
\fontsmall
\begin{aligned}
\alpha^{*}_{i} = argmax_{\alpha} log \Gamma(\sum_{l} \alpha_{l}) - \sum_{l} log \Gamma(\alpha_{l}) \\ + \sum_{l}\left((\alpha_{l} - 1) (log \hat{x_{l}}) \right), 
\text{where  } \hat{x_{l}} = \frac{1}{T} \sum_{i} log(x_{i_{l}})
\end{aligned}
\end{equation}

Then, the log likelihood given distribution $Q_{0}$ is calculated using equation \ref{log} below:
\vspace{-.1in}
\begin{equation} \label{log}
\fontsmall
\begin{aligned}
LL(x_{1} \dots x_{T}, Q) = \sum^{T}_{i=1} log(Q(x_{i}))  \\
Q(x_{i}) = \frac{1}{B(\alpha)} \prod^{d}_{l = 1} x^{\alpha_{l}-1}_{i_{l}}, \text{ and }
B(\alpha) = \frac{\prod_{l=1}^{d} \Gamma(\alpha_{l})}{\Gamma(\sum_{l=1}^{d} \alpha_{l})}\\
\text{where   } d = |x_{i}| \text{ Dimensionality of } x_{i}, \text{  } x_{l} \geq 0\text{, and} \sum_{l=1}^{d} x_{l} = 1. 
\end{aligned}
\end{equation}

Then, at each time step $t$, that is seen as a potential change point, we split the data into two parts (prior and after this time step $t$), and we estimate the maximum likelihood as well as the sum of log likelihood for both partitions using the equations above. Finally, the algorithm returns the point in time $T^{*}$ associated with the maximum log likelihood to be a potential change point. If the difference between this value and the log likelihood of our unsplit original data turns out to be greater than our threshold, then we declare that a change has been detected at time $T^{*}$.

\begin{algorithm}[!htb]
	\caption{ \small Dirichlet Change Point Detection Algorithm} \label {sim}
	\begin{minipage}{\linewidth}
		\fontsmall
		\begin{algorithmic}[1]
			\State \textbf{Input} Time Window [$1 \dots T$], Data [$x_{1} \dots x_{T}$].
			\State \textbf{Output} $T^{*}$: Change Point (if there is a change).
			\Procedure{DCP}{[$x_{1} \dots x_{T}$]}
			\State $Q_{0} \gets$ \textbf{Estimate} Drichlet Parameters for [$x_{1} \dots x_{T}$] using Eq. \ref{drichlet}
			\State $LL_{0} \gets$ \textbf{Estimate} Log-Likelihood for [$x_{1} \dots x_{T}$] under $Q_{0}$ (Eq. \ref{log}).
			\State $(T^{*}, LL^{*}) \gets$ \textsc{estimate\_2window}([$x_{1} \dots x_{T}$]) 
			\State $Z^{*} \gets LL^{*} - LL_{0}$
			\If{$Z^{*} > \text{threshold}$}
				\State \textbf{Return} Change point at $T^{*}$.
			\Else
				\State No change, \textbf{Return}
			\EndIf
			\EndProcedure
			\Procedure{estimate\_2window}{[$x_{1} \dots x_{T}$]}
			\For {$t \in 1 \dots T-1$}
				\State $Q_{1} \gets$ \textbf{Estimate} Drichlet Parameters for [$x_{1} \dots x_{t}$] (Eq. \ref{drichlet}).
				\State $Q_{2} \gets$ \textbf{Estimate} Drichlet Parameters for [$x_{t+1} \dots x_{T}$] (Eq. \ref{drichlet}).
				\State $LL_{t} \gets$ \textbf Log-Likelihood for [$x_{1} \dots x_{t}$] under $Q_{1}$ + Log-Likelihood for [$x_{t+1} \dots x_{T}$] under $Q_{2}$ (Eq. \ref{log}).
			\EndFor
			\State $LL^{*} \gets max_{(t \in 1 \dots T-1)} LL(t)$
			\State $T^{*} \gets argmax_{(t \in 1 \dots T-1)} LL(t)$
			\State \textbf{Return} $(T^{*}, LL^{*}) $
			\EndProcedure
		\end{algorithmic} 
	\end{minipage}
\end{algorithm}


\begin{figure*}
	\begin{center}
\includegraphics[trim=50 0 50 40, width=0.93\textwidth]{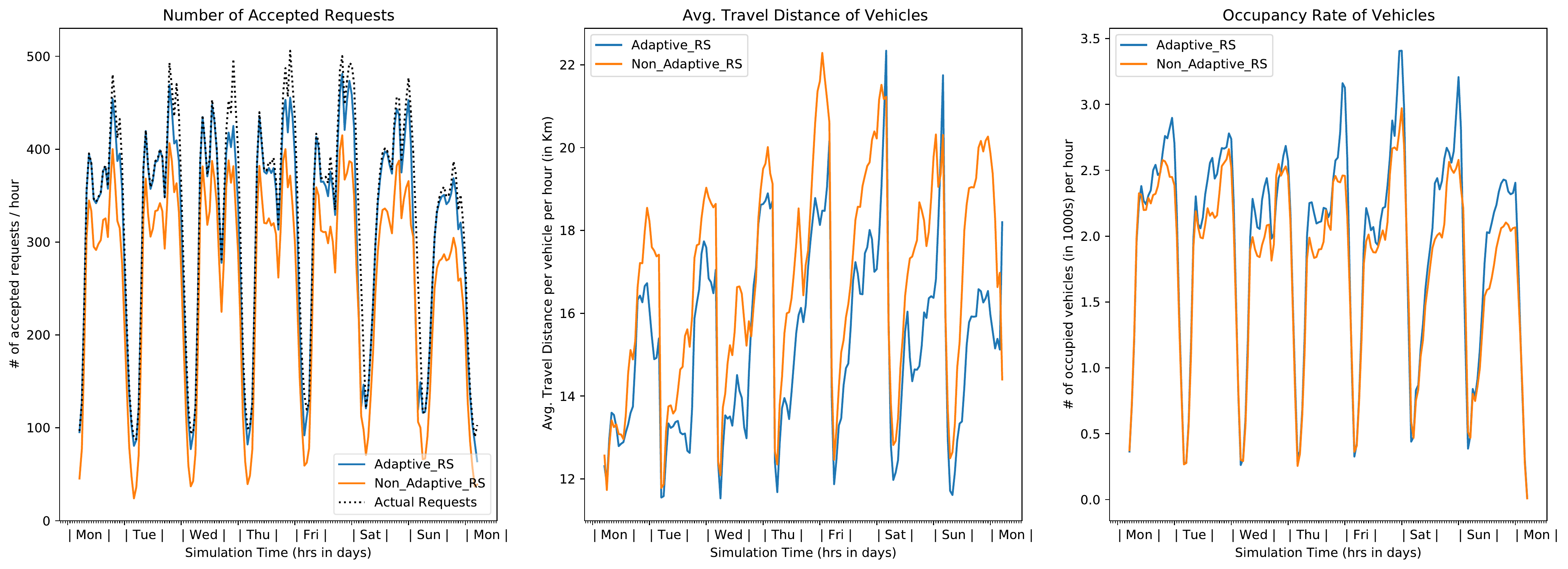} 
\vspace{-0.08in}
\caption{Evaluation Metrics for AdaPool and the non-adaptive baseline}\label{result}
\end{center}
 \vspace{-0.13in}
\end{figure*}


\vspace{-1em}
\subsection{DQN Dispatch Policy} \label{dispatch}
In this section, we detail our system's global reward objective which allows efficient fleet dispatch in fulfilling service workloads. This global reward is optimized by our proposed algorithm in a distributed fashion as vehicles solve their own DQN to maximize rewards. At each time step $t$, each vehicle that is either idle or newly entered the market (i.e. vehicles that are marked available $ \forall V_{j} \in v_{t,m}$ ) needs to make a dispatch decision of which zone $m$ to be dispatched to at time slot $t$. To take this decision, each vehicle calculates the discounted reward (Q-value) associated with each potential action and picks the action that would yield the maximum future reward. The reward function, which drives the dispatch policy learner's objectives, is shaped in a manner which aims to (1) satisfy the demand of pick-up orders, thereby minimize the supply-demand mismatch: ($\text{diff}_{t}$), (2) minimize the dispatch time: $T_{t}^{D}$ (i.e.,  expected travel time of vehicle $V_{j}$ to go zone $m$ at time $t$), (3) minimize the extra travel time a vehicle takes for car-pooling compared to serving one customer: $\Delta t$, (4) maximize the fleet profits $P_{t}$, and (5) minimize the number of utilized vehicles: $e_{t}$.  These objectives are defined in Appendix \ref{apdx:reward}.

The overall objective of the system is optimized at each vehicle in the distributed transportation network. In this case, the reward $r_{t,n}$ for vehicle $n$ at time slot $t$ is represented in Eq. \eqref{individual}, where the objectives above are mapped to: (1) $C_{t,n}$: number of customers served by vehicle $n$ at time $t$, (2) dispatch time: $T_{t,n}^{D}$, (3) extra travel time: $T_{t,n}^{E}$, (4) average profit for vehicle $n$ at time $t$: $\mathbb{P}_{t,n}$, and (5) $\max (e_{t,n} - e_{t-1,n}, 0)$ that addresses the objective of minimizing the the number of vehicles at $t$ to improve vehicle utilization.   The reward function of each vehicle is defined as a weighted sum of these terms as:
\begin{equation} \label{individual}
\begin{multlined}
r_{t, n} = r(s_{t,n}, a_{t,n})
= \beta_{1} C_{t,n} - \left[ \beta_{2} T_{t,n}^{D} + \beta_{3} T_{t,n}^{E} \right] + \\ \beta_{4} \mathbb{P}_{t,n} - \beta_{5} [\text{max}(e_{t,n} - e_{t-1,n}, 0)]
\end{multlined}
\end{equation}

where $\beta_{1}, \beta_{2}, \beta_{3}, \beta_{4} \text{ and } \beta_{5}$ depend on the weight factors of each of the objectives. Note that we maximize the discounted reward over a time frame. The negative sign here indicates that we want to minimize these terms.  Further, the last term captures the status of vehicle $n$ where $e_{t,n}$ is set to 1 if vehicle $n$ was empty and then becomes occupied at time $t$ (even if by one passenger), however, if it was already occupied and just takes a new customer, $e_{t,n}$ is 0. The intuition here is that if an already occupied vehicle serves a new user, the congestion and fuel costs will be less when compared to when an empty vehicle serves that user. Note that if we make $\beta_{3}$ very large, it will dis-incentivize passengers and drivers from making detours to serve other passengers, Thus, the setting becomes similar to the one in \cite{fleet_oda}, where there is no carpooling. The overall optimization process includes a route planning and matching policy, and the DQN dispatch policy working in tandem with each other.

While the primary role of the DQN is to act as a means of dispatching idle vehicles, it contains useful signals on future anticipated demand that is utilized by other components of our method including the Demand Aware Matching and Route Planning. The additional profits term $P_{t}$ integrated with the reward function makes the output expected discounted rewards (Q-values) associated with each possible move on the map, a good reflection of the expected earnings gained when heading to these locations. The Q-values are then used to decide on the best dispatching action to take for each individual vehicle. Since the state space is large, we don't use the full representation of $s_{t}$, instead a map-based input is used to alleviate this massive computing.

\section{Experimental Results} \label{results}
\subsection{Simulator Setup}
In our simulator, we used the road network of the New York Metropolitan area along with a real public dataset of taxi trips in NY \cite{10}.  We used \textit{Python} and \textit{Tensorflow} to implement our framework.  For each trip, we obtain the pick-up time, passenger count, origin location,  drop-off location and ride fare. We use this trip information to construct travel requests and demand prediction model. We start by populating vehicles over the city, randomly assigning each vehicle a type and an initial location. According to the type assigned to each vehicle, we set the accompanying features accordingly such as: maximum capacity, mileage, and price rates (per mile of travel distance $\omega^{1}$, and per waiting minute $\omega^{3}$). We initialize the number of vehicles, to $8000$. Note that, not all vehicles are populated at once, they are deployed incrementally into the market by each time step $t$.  We also defined a reject radius threshold for a customer request. Specifically, if there is no vehicle within a radius of 5km to serve a request, it is rejected. This simulator hosts each deep reinforcement learning agent which acts as a ridesharing vehicle that aims to maximize its reward: Eq. \eqref{individual}.

\vspace{-0.1in}
\subsection{DQN Training and Testing}
The fleet of autonomous vehicles was trained in a virtual environment that simulates urban traffic.  We consider the data of June 2016 for training and one week from July 2016 for evaluations. For each experiment, we trained our DQN neural networks using the data from the month of June 2016 for $20k$ epochs, which corresponds to a total of 14 days, and used the most recent 5000 experiences as a replay memory.  Upon saving Q-network weights, after training, we retrieve the weights to run testing on an additional 8 days from the month of July which corresponds to $10k$ epochs. Thus, T = 8 $\times$ 24 $\times$ 60 steps, where $\Delta t$ = 1 minute. 
To initialize the environment, we run the simulation for 20 minutes without dispatching the vehicles. Finally, we set $\beta_{1} = 10, \beta_{2} = 1, \beta_{3} = 5$, $\beta_{4} = 12$, $\beta_{5} = 8$. Each vehicle has a maximum working time of $21$ hours per day, after which it exits the market.  Also, we perform hyper-parameter tuning to set $k$ (the number of models to be learnt by our DQN) to 7, and the threshold for our ODCP algorithm to $5000$. We show that our framework is able to recognize up to 7 different diurnal patterns throughout the day.

\begin{figure}
\centering
\includegraphics[trim = 50 0 40 10, width=.37\textwidth]{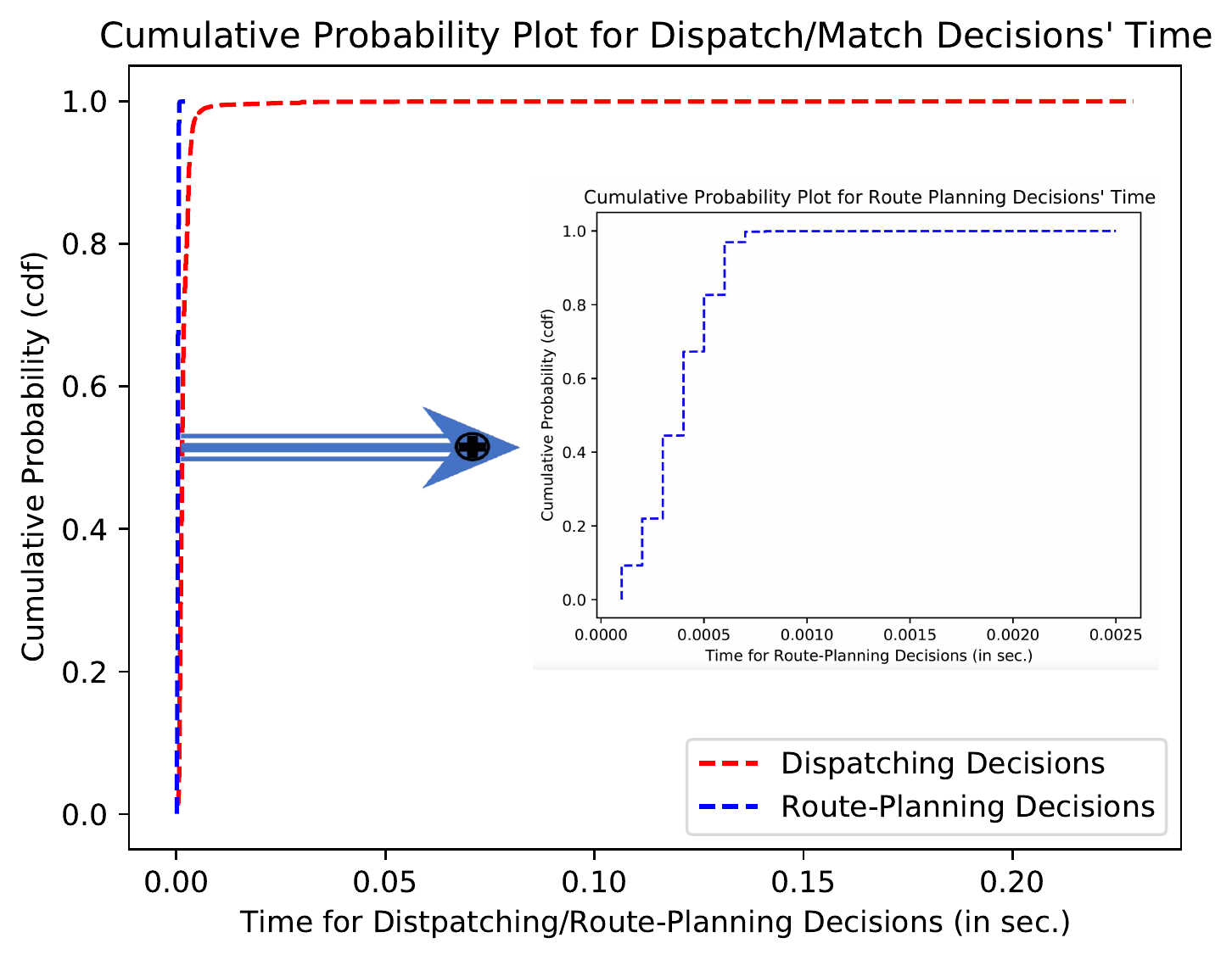}
\caption{\small CDF for Dispatching and Route-Planning Decisions' Times} \label{dis_match_time}
\vspace{-1em}
\end{figure}

\begin{figure*}
  \centering
 \begin{tabular}{cc}
 \begin{subfigure}{0.48\textwidth}
\includegraphics[trim=30 10 0 5, width=\textwidth]{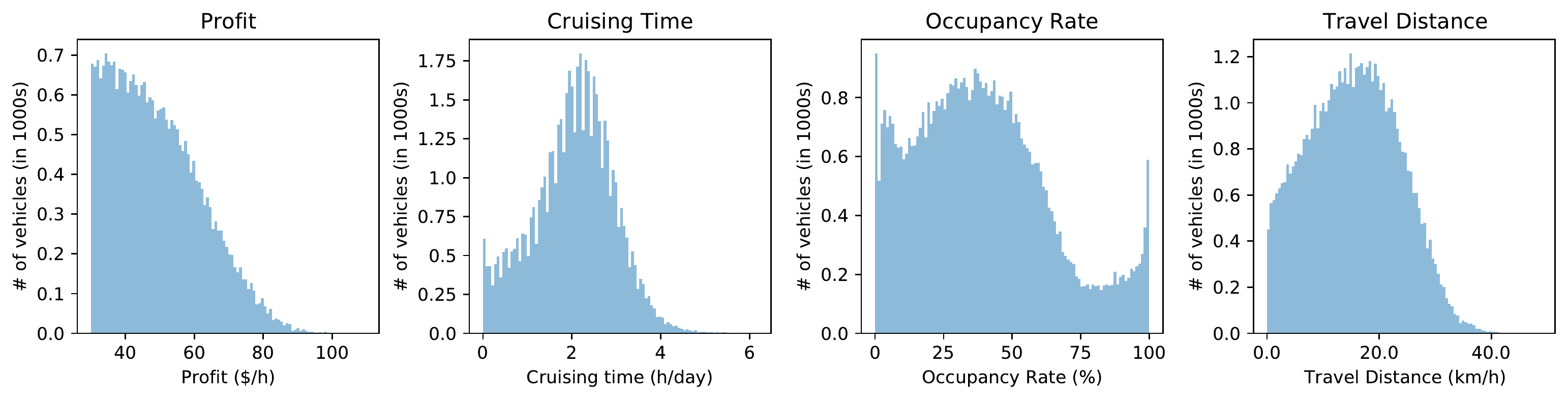} 
\vspace{-0.1in}
\caption{\small Performance Metrics of our AdaPool Framework}\label{ada_sumary}
\end{subfigure} \hspace{1em}
\begin{subfigure}{0.48\textwidth}
\includegraphics[trim=10 10 30 5, width=\textwidth]{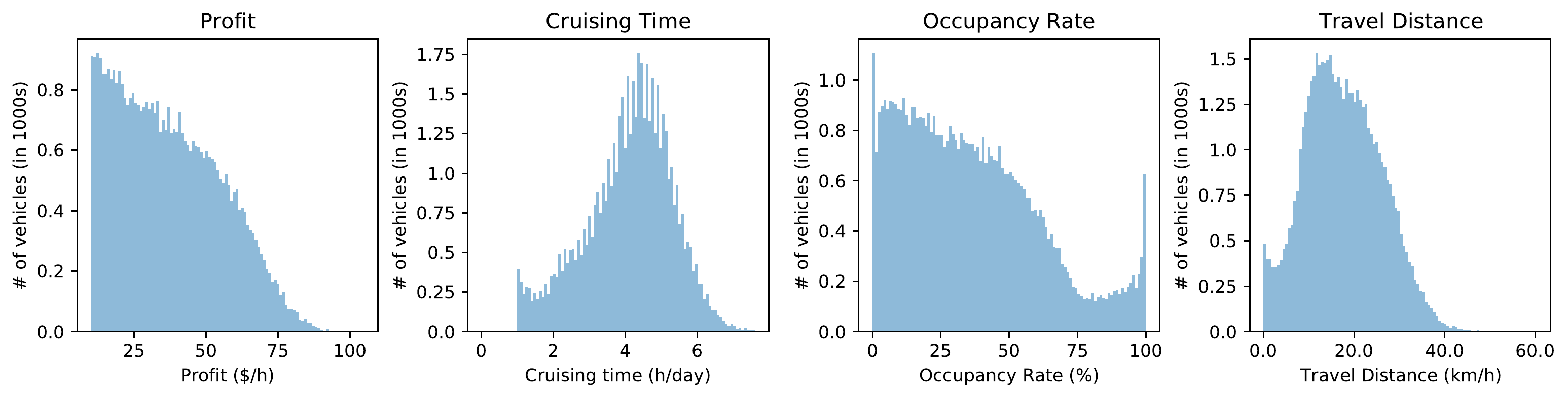} 
\vspace{-0.1in}
\caption{\small Performance Metrics of the Non-Adaptive Baseline}\label{non_sumary}
\end{subfigure}  \\ [1em]
\end{tabular}
 \caption{Histograms of Performance Metrics for the Proposed AdaPool and the Baseline} \label{tab1}
 \vspace{-0.1in}
\end{figure*}

\vspace{-0.1in}
\subsection{Computational Analysis}
To provide more insight regarding the complexity of our AdaPool framework, we investigate:

{\bf Dispatch Decisions: } We show in Fig. \ref{dis_match_time}, the cumulative distribution function (cdf) plot for the time taken for dispatch decisions for each individual vehicle. We can obseve that with probability 1.0, it will take the vehicle $< 0.2$ seconds to make a dispatch decision of which location on the map to head to next in order to maximize its own reward.

{\bf Matching and Route-Planning Decisions: } We show in Fig. \ref{dis_match_time} the cdf plot for the time taken for route-planning decisions for each individual vehicle.  This is the time taken by the vehicle to apply the greedy insertion operations and decide on the $k^{\prime}$ ride requests, which satisfy its capacity constraint and correspond to the minimum cost. We can conclude that with probability 1.0, it will take the vehicle $< 2.5$ milliseconds to make a dispatch decisions. 

These results proves the viability and efficiency of our framework to be applied in large-scale real-world environments.  {Clearly, the time taken for route-planning is negligible when compared to the time taken for making dispatching decisions; however, both are very reasonable for real-time decision making scenarios}.

\vspace{-0.1in}
\subsection{Performance Metrics}
We breakdown the reward, and investigate the performance of AdaPool against a non-adaptive baseline \cite{darm}. Recall that we want to minimize the components of our reward: Eq. \eqref{individual}.
\begin{itemize}[leftmargin = *]
\item \textbf{Accepted Requests}:  {This is calculated as the total number of requests served by the fleet per working hour. }The total number of customers served indicates how effectively the algorithm is able to minimize the supply demand gap and fulfill delivery requests. 
\item \textbf{Travel Distance:} This metric shows the total amount of distance traveled by each vehicle per hour of service, which gives a good reflection of the cost incurred by vehicles due to serving multiple ride requests.   {This distance is computed using the weights of the $n$ edges that constitute the vehicle’s optimal route from its current location to origin $o_{i}$ and to destination $d_{i}$. This route is obtained through the insertion-operation, the route which minimizes the DARM cost function as shown in Algorithm \ref{insert}.}
\item \textbf{Occupancy Rate:} This metric captures the utilization rate of the fleet of vehicles, it keeps track of how many vehicles are deployed from the fleet to serve the demand.   {This is calculated as the total number of vehicles that are carrying passengers per hour of service, while we also calculate the occupancy rate of vehicles (in Fig.  \ref{tab1}), which is defined as the percentage of time where vehicles are occupied out of their total working time.} By minimizing the number of occupied vehicles, we achieve better utilization of individual vehicles in serving the demand. A lower occupancy rate indicates that a fleet is able to minimize the number of vehicles on the street to serve the requests. 
\item \textbf{Profit:} This represents the net profit accumulated by a vehicle over the course of a day,   {where the cost incurred by fuel consumption is subtracted from the revenue.  The revenue is calculated by summing the trip fares from all customers served by this vehicle}.
\item \textbf{Cruising (idle) time:} This represents the time during which a vehicle is neither occupied nor gaining profit but still incurring gasoline cost.  Lower cruising times therefore suggests a cost effective policy. 
\item \textbf{Detected points of change and the corresponding demand and hour in day: } This measure helps investigate the pattern of change in demand against the detected points of change in order to validate if our framework adapts accordingly.
\end{itemize}
The proposed non-adaptive baseline aims to evaluate the effectiveness of the adaptive aspect of our framework as well as the demand aware matching and route planning component.  Our proposed method incorporates both insertion-based route-planning and diurnal pattern adaptation. As compared to the non-adaptive baseline, we hypothesize that our AdaPool framework would be a more effective approach.  With the capability of adapting to the demand pattern, we expect AdaPool to bring the supply/demand gap to a minimum and thus, minimize the cruising idle time and travel distance in addition to increasing the overall accept rate of requests. Given that the core intuition of our matching and route planning components is to group together rides that share route intersections to their destinations as opposed to rides heading to opposite-direction-destinations, we expect improvements in the number of rides served, profits,  travel distance, and occupancy rate.

\vspace{-0.1in}
\subsection{Results Discussion}
From our simulation, we observe that the hypothesis for our baseline comparison has been supported for the most part by our experimental results.  In Figure \ref{result}, we investigate the overall performance of our AdaPool framework. We show the actual number of requests as the dotted black line.  We can observe that, over a week long of simulation, AdaPool consistently improves the overall acceptance rate of ride requests by around a $10 - 15 \%$, while at the same time, significantly decreases the average travelling distance of the fleet. This proves the effectiveness of AdaPool in minimizing the supply/demand mismatch. Although, this comes at the cost of a slight increase in the number of utilized vehicles ($\approx$ 300 extra vehicles) in the fleet, this outcome proves that - unlike the non-adaptive baseline, AdaPool does a successful job in re-balancing vehicles over the city, so extra vehicles would become occupied in order to serve extra demand (that was not served in the non-adaptive scenario) while at the same time achieving a decrease in the average travel distance of the fleet. This result points towards the effeciency of our insertion-based route-planning as it allows for serving more requests with the smallest possible travel distance. It is also worth noting that, even with a slight increase in the number of occupied vehicles, the total number of utilized vehicles stays below $3.5k$ which is less than half of the fleet (we set the maximum number of vehicles to $8000$ in our simulator). This proves our hypothesis that AdaPool outperforms the non-adaptive baseline in the utilization of available resources. This is a positive outcome that points towards the viability of our proposed approach to learn diurnal patterns and adapt in a timely manner.

\begin{figure}
  \centering
 \begin{subfigure}{0.24\textwidth}
\includegraphics[trim=50 0 20 40, width=\textwidth]{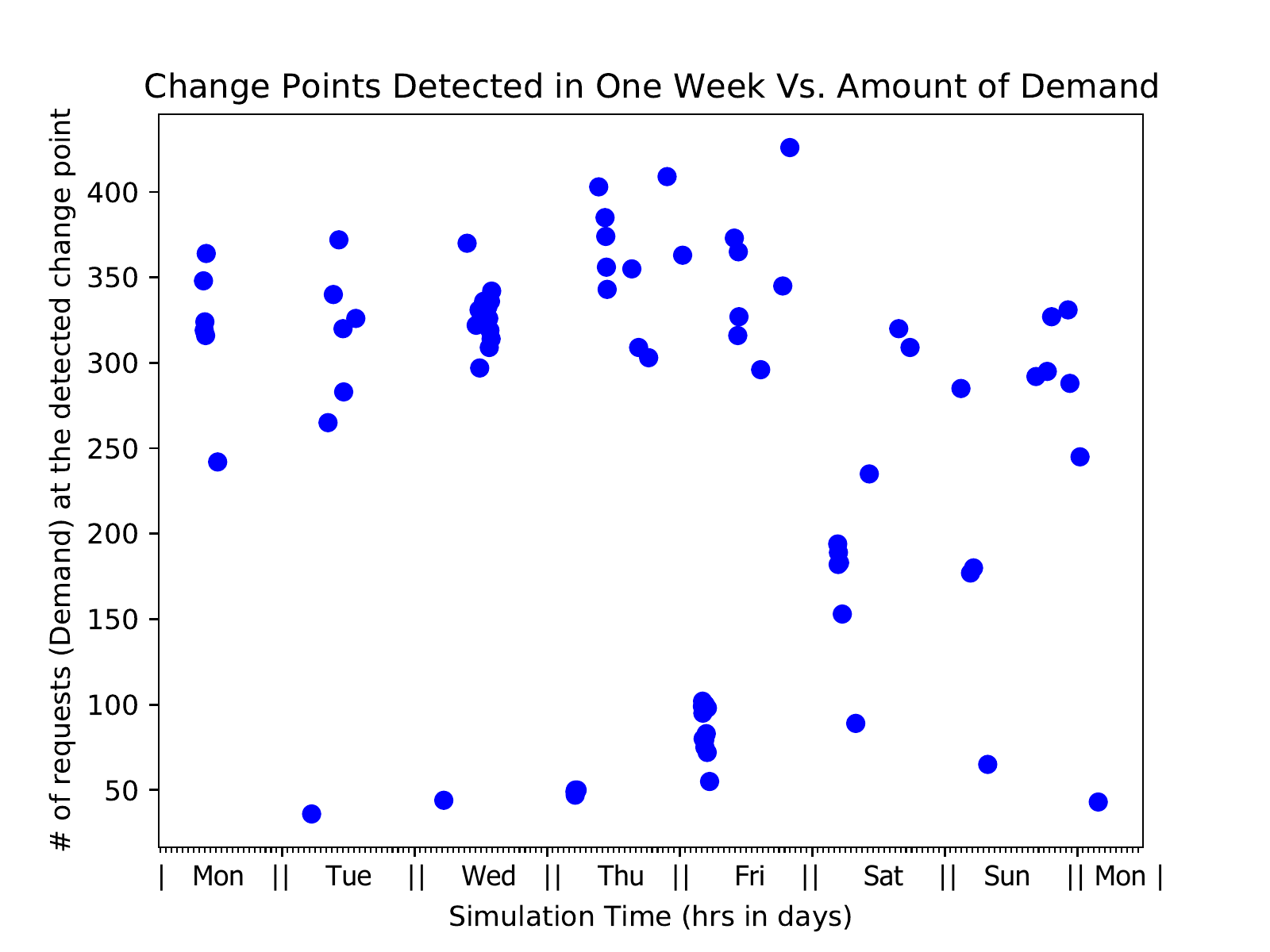} 
 \vspace{-0.08in}
\caption{Detected change points and their corresponding demand}\label{switch}
\end{subfigure} 
\begin{subfigure}{0.24\textwidth}
\includegraphics[trim=20 0 50 40, width=\textwidth]{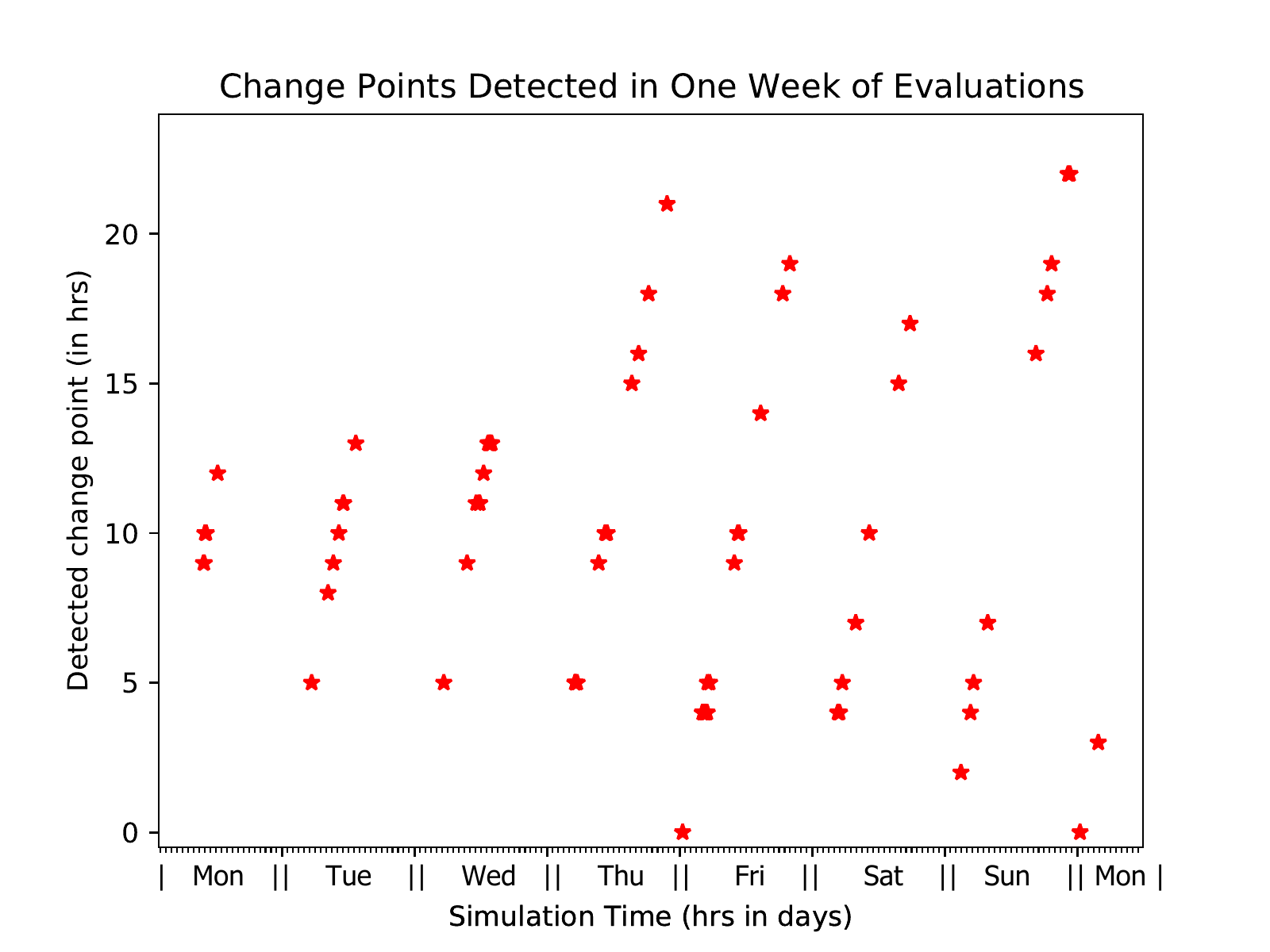}
 \vspace{-0.08in}
\caption{Detected change points and their corresponding hour in day}\label{switch_hr}
\end{subfigure} 
 \caption{Analysis for Diurnal Pattern Adaptation}
 \vspace{-0.1in}
\end{figure}

Besides, the extra number of utilized vehicles (which can be explained by the additional number of requests served) will -in turn- increase the average profits of the fleet which is also another desirable outcome. In Figure \ref{ada_sumary}, we can observe that AdaPool increases the average profits per vehicle in the fleet while minimizing their cruising idle time, which in turn, cuts down fuel consumption and environmental pollution. We  note that, in AdaPool, more than $80\%$ of vehicles in the fleet spend less than 3 hours  of idle cruising per day. On the other hand, with the non-adaptive scenario, the plot is more skewed to the right where the average idle cruising time reaches more than 5 - 6 hours per day. This is mainly due to AdaPool dispatching vehicles according to the learnt demand pattern which makes the vehicles present at locations very close to the anticipated demand and thus minimizes their idle time and  cruising time to get to the requests assigned to them.  On the other hand, the non-adaptive baseline relies only on mobilizing vehicles according to the pickup locations of their requests as opposed to learning the changes of the supply-demand distribution of the city and mobilizing accordingly when they experience idle time. 


In Figure \ref{result}, we observed the occupancy rate of the fleet (how many vehicles are occupied from the fleet); however in Figure \ref{tab1}, we look at the occupancy rate per vehicle. That is how much of the time, when the vehicle is in service, does it stay occupied. We note that, with AdaPool, more than half of the fleet is occupied for more than $50\%$ of their time in the market. Again, the occupancy rate plot is more skewed to the left with the non-adaptive baseline, which suggests less occupancy rate per vehicle as it leans towards $10\% - 30\%$. Finally, after we investigated the overall average travel distance for the whole fleet in Figure \ref{result}, we also take a closer look at the travel distance per vehicle in Figure \ref{tab1}. Clearly, AdaPool shows a considerably less travel distance per vehicle than the non-adaptive baseline. This further ascertains that, in AdaPool, the insertion-based route-planning in tandem with the demand pattern adaptaion achieves a great success in learning the demand pattern and re-balancing vehicles over the city accordingly. This significantly improves the utilization of each individual vehicle as well as the whole fleet. 

Finally, to validate the adaptation aspect of AdaPool, Figure \ref{switch} shows the amount of demands at the detected points of change at which our approach switches contexts (i.e., models). We can, clearly, conclude that the detected change points are associated with either a sharp increase or decrease in demand which suggests a change in context; thus, our AdaPool switches to either learn a new model or to a previously learnt model (if the samples it observes has been learnt before). This behavior is consistent throughout the week of testing, except over the weekend where we can observe there has been some detected change points at intermediate demand. This could be attributed to other factors represented in our reward function, where the change might not be only dependent on the amount of demand but also on factors resulting from that such as larger customer's waiting times or larger cost incurred from fuel consumption. For future work, we plan to conduct more investigation on weekends, holidays, etc. and research on what other aspects could attribute to changing contexts (such as: unexpected weather conditions, etc.).

In Figure \ref{switch_hr}, we take a closer look at the pattern of the detected changes. We can observe that the change pattern is relatively consistent throughout the week of evaluations, excluding the weekend where the pattern varies slightly. In weekdays, AdaPool detects changes somewhere between 5-7 am, then between 11-noon, after that between 4-6 pm, and finally at night between 8-9 pm. On the other hand, for weekends there are additional detected points later in the night around 10-11 pm and at mid-night. Tying back to Figure \ref{switch}, we can conclude that these points correspond to peaks (e.g. 4-6 pm when people heading home from work, and students from school), decreasing demand (e.g., around 8-9 pm when traffic generally starts slowing down in weekdays), or rising peaks (e.g., around 5-7 am when people starts heading to work, or to school). This outcome ascertains that our approach is able to detect diurnal patterns as the contexts of the underlying environment change. In addition to this, AdaPool also adapts according to these changes, as it switches between models. Tracking these model switches, we observe that AdaPool exploits the same model around the same timing each day, and thus, models vary by the variation of demand as well. For instance, in weekdays, between 10 pm and 5 am, AdaPool utilizes the same model every day. We note that this time-frame characterizes the least amount of demand every weekday. Similarly, mornings between 5 - 9 am, the same model is also utilized every week day. This time-frame corresponds to a notably high demand as it signals the beginning of each business day. Therefore, AdaPool is able to adapt by switching to the model that corresponds to the diurnal pattern it has learnt, while it still is able to learn online any new unexpected changes as they take place within the environment.
\vspace{-0.1in}
\section{Conclusion and Future Work} \label{conc}
In this paper, we detailed our novel approach\textemdash Distributed Adaptive Deep Q Learning for Ride-Sharing with car pooling, namely ``AdaPool'' framework and our Demand-Aware Matching and route planning approach\textemdash that generate ideal routes on-the-fly and adapts dynamically to changing environmental contexts. Agents' (i.e., vehicles) decision-making process is informed by a reward function that aims to achieve the maximum profit for drivers while accounting for fuel costs, waiting times, and supply-demand mismatch to compute the reward. This novel AdaPool methodology integrates a DQN-based dispatch algorithm, learns up to 7 different contextual models, and adapts accordingly by detecting the relevant change points. The learnt Q-values associated with the active model are then leveraged by each vehicle to make informed dispatch decisions independently. 
Experimental results show that AdaPool framework boosts the acceptance rate, while enhancing drivers' profits and decreasing their average travel distance. Given the maximum number of vehicles (8000) populated in the simulation, our framework utilizes less than 50\% of the vehicles to serve the demand of up to 90\% of the requests. Experiments also show that vehicle idle time (cruising without passengers) is reduced to below 2 hours per day and 40\% - 70\% of the vehicles are occupied most of the time. Our model-free AdaPool framework can be extended to large-scale ridesharing protocols due to the vehicles' distributed decision making that reduces the decision space significantly. \\
Application of the proposed approach for decision making in other environments with diurnal patterns will be considered in the future.  Extension of this work to include capabilities of a joint delivery system for passengers and goods as in \cite{manchella2020flexpool,manchella2020passgoodpool},  {multitrip deliveries within a certain time window as in \cite{wang2020solving},} or using multi-hop routing of passengers as in \cite{singh2019distributed} 
for efficient fleet utilization is left as future work.

\bibliographystyle{IEEEtran}
\bibliography{IEEEabrv,IEEEexample}
\newpage

\appendices
\section{Notations}\label{apd:notations}
The key notations used in this paper are summarized in Table \ref{notations}.
\begin{table}
	\centering
	{
			\resizebox{.96\columnwidth}{!}{%
		\begin{tabular}{||c|c||}
			\hline
			\textbf{Parameters} & \textbf{Description}\\
			\hline
			$N$ & The number of vehicles.\\
			\hline
			$M$ & The number of regions.\\
			\hline
			$\Delta t$ & Step Size.\\
			\hline
			$T$ & Maximum number of time steps.\\
			\hline
			$s_{t}$ & The state of the environment at time $t$.\\
			\hline
			$a_{t}$ & The dispatch action taken at time $t$. \\
			\hline
			$\eta \in [0, 1]$ & Time discount factor.\\
			\hline
			$r_{t}$ & The reward gained at time $t$. \\
			\hline
			$d_{t,m}$ & The number of predicted requests for zone $m$ at time $t$.\\
			\hline
			$D_{t:T}$ & \thead{The future predicted demand from \\time $t$ to time $t+T$ over all regions.}\\
			\hline
			$d_{t,\tilde{t}, m}$ & \thead{The number of vehicles that will become \\ available at time $\tilde{t}$ at region $m$.}\\
			\hline
			$V_{t:t+T}$ & \thead{The predicted supply (number of available vehicles)\\ from time $t$ to time $t+T$ over all regions.}\\
			\hline
			$X_{t}$ & State vector that hold the $N$ vehicles' status at time $t$.\\
			\hline
			$S_{V_{j}}$ & The current route of Vehicle $j$.\\
			\hline
			$S^{\prime}_{V_{j}}$ & \thead{The route with minimum insertion cost \\ after route-planning for vehicle $j$}.\\
			\hline
			$k$ & Number of initial requests to be assigned to the vehicle.\\
			\hline
			$k^{\prime}$ & Number of final requests to be served by the vehicle.\\
			\hline
			$(o_{i}, d_{i}) \in r_{i}$ & \thead{Origin (pickup) and destination (dropoff)\\ locations associated with the $i^{th}$ request.}\\
			\hline
			$\mathbb{P}_{t,n}$ & Average Profit for vehicle $n$ at time $t$.\\
			\hline
			$C_{t,n}$ & The number of customers served by vehicle $n$ at time $t$.\\
			\hline
			$T_{t,n}^{D}$ & Time taken for dispatch by vehicle $n$ at time $t$.\\
			\hline
			$T_{t,n}^{E} $ & Extra Travel Time by vehicle $n$ at time $t$.\\
			\hline
			$M_{c}$ & The current active model.\\
			\hline
			$\theta$ & The network parameters in Q-network (DQN).\\
			\hline
		\end{tabular}
	}
		\caption{Notations used in AdaPool Formulations}\label{notations}
	}
\end{table}

\section{Control Unit Components:}\label{models}
\subsection{OSRM and ETA Model}
We construct a region graph relying on the New York city map, obtained from OpenStreetMap \cite{osm}. Also, we construct a directed graph as the road network by partitioning the city into small service area 212 x 219 bin locations of size 150m x 150m. We find the closest edge-nodes to the source and destination and then search for the shortest path between them. To estimate the minimal travel time for every pair of nodes, we need to find the travel time between every two nodes/locations on the graph. To learn that, we build a fully connected neural network using historical trip data as an input and the travel time as output. The fully connected multi-layer perception network consists of two hidden layers with width of 64 units and rectifier nonlinearity. The output of this neural network gives the expected time between zones. While this model is relatively simple (contains only two hidden layers), our goal is to achieve a reasonable accurate estimation of dispatch times, with short running time. Finally, if there is no vehicle in the range of $5 \; km^{2}$, the request is considered rejected.

In the ETA model, we want to predict the expected travel time between two zones (two pairs of latitudes and longitudes). We split our data into 70\% train and 30\% test. We use day of week, latitude, longitude and time of days as the explanatory variables and use random forest to predict the ETA. The final ETA model yielded a root mean squared error (RMSE) of 3.4 on the test data.

\subsection{Demand Prediction Model}
\begin{figure}
	\vspace{-2.5em}
	\begin{center}
		\includegraphics[width=0.5\textwidth]{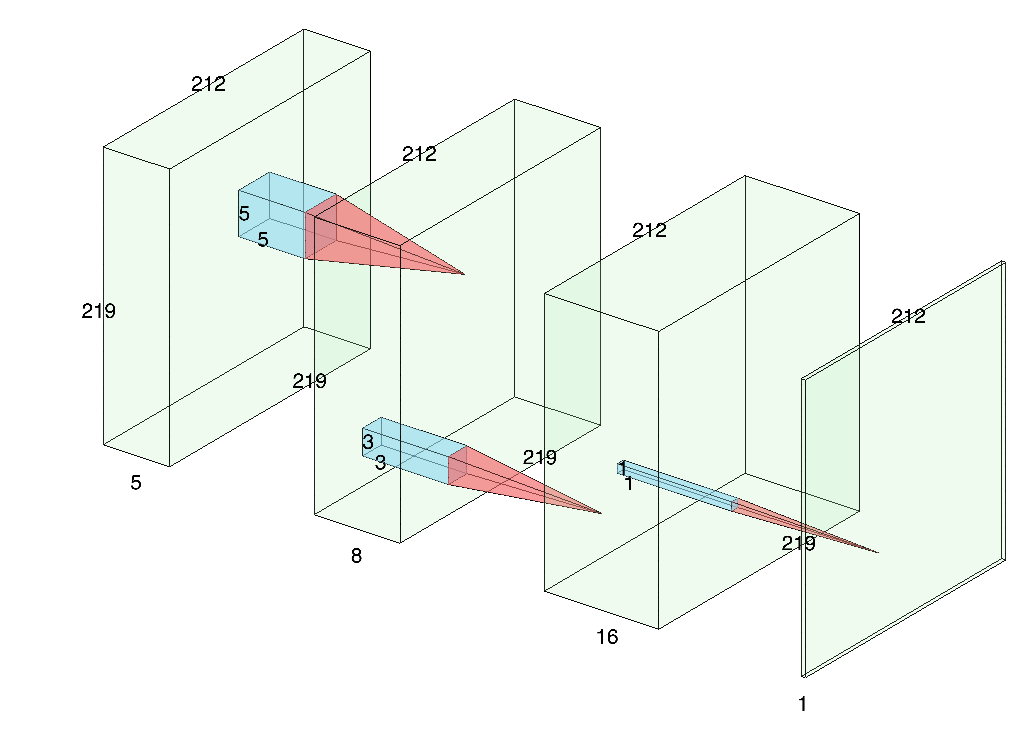}
		\vspace{-3em}
		\caption{The architecture of the Convolution Neural Net used for Demand Prediction.} \label{demand}
	\end{center}
\end{figure}

The demand prediction model is a critical element to the simulator in building the state space vector that allows DQN agents to proactively dispatch towards areas where there is a high demand. We use a convolutional neural network to predict future demand. Figure \ref{demand} shows the architecture of this Conv-Net. The output of the network is a $212 x 219$ image such that each pixel represents the expected number of ride requests in a given zone for $30$ minutes ahead. The network input consists of two planes of actual demand of the last two steps. The size of each plane is 212 x 219. The first hidden layer convolves $16$ filters of size 5 x 5 with a rectifier nonlinearity, while the second layer convolves 32 filters of 3 x 3 with a rectifier nonlinearity. The output layer convolves 1 filter of size 1 x 1 followed by a rectifier nonlinear function.

\begin{figure}
	\captionsetup{justification=centering, font=small}
	\centering
	\includegraphics[trim = 70 50 50 70, width=.4\textwidth]{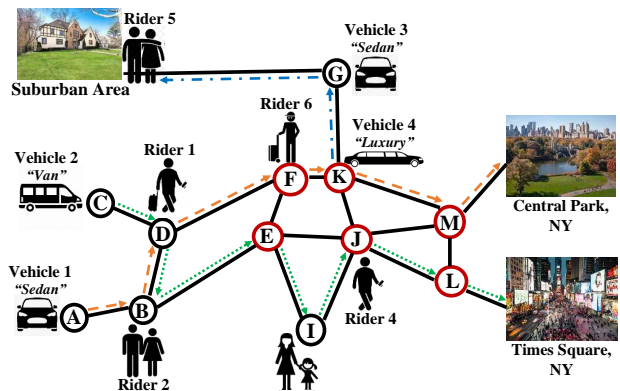}
	\vspace{-1em}
	\caption{\small A diagram to illustrate routing in a region graph, with 13 regions, A to M. The orange dotted line shows the path of Vehicle 1, the green dotted line for Vehicle 2, and the blue for vehicle 4 in the distributed adaptive ride-sharing scenario.} \label{scenario}
	\vspace{-1em}
\end{figure}

\vspace{-0.12in}
\section{{Dynamic Ridesharing Example:}}
{In Fig. \ref{scenario}, we consider a situation that highlights the differences between various ride-sharing scenarios. The figure demonstrates the locations of passengers and vehicles at a given point in time. Riders 1 and 6 are heading to Central Park, NY, while riders 2, 3, and 4 want to go to Times Square, NY. Rider 5 wants to be picked up from a suburb area that is more than 10 - 15 miles away from the downtown area. The regions circled in red represent the downtown area where a high demand is anticipated. A basic scenario is for vehicle 2 to serve rider 1, vehicle 1 to serve rider 2, vehicle 3 to serve rider 5 and vehicle 4 to serve rider 6. In this scenario, no ride-sharing is involved and no consideration of vehicles' types or customers'/drivers' preferences takes place. Thus, all four vehicles are used to serve only part of the demand. Since no more vehicles are available, both rider 3 and 4 are left out. }
 {Now, we consider a ride-sharing scenario where vehicle 2 serves both rider 1 and 6 (due to its proximity to rider 1), similarly vehicle 1 serves both rider 2 and 3. Rider 4 will still be left out due to the capacity limit on vehicle 1. Rider 4 can now be served via vehicle 4, but that is efficient neither for the driver (as he travels large idle distance) nor for the customer (as the luxurious vehicle would cost much more, even though it might not be the customer's prefereference but it is the only option available). Finally, vehicle 3 serves rider 5. In this scenario, ride-sharing allows a higher accept rate where all requests are accommodated. However, this might not be the most efficient vehicle-customer matching in terms of ride fares, and how much profit drivers make given that they might need to travel large idle distances to reach customers. Thus,  an adaptive approach that can learn the underlying demand patterns while optimizing for the overall fleet objectives is essential for the dynamic ridhesharing environment.}

{Using our algorithm, we anticipate the demand in the route to Times Square to be very high (particularly on weekends) by analyzing the historical distribution of trips across zones. So sending a van down this road is a smart decision that is made by our algorithm to have enough capacity to accommodate as many requests as possible that are located along the way to the destination. Therefore, vehicle 2 serves riders 2, 3 and 4 (see green line). Rider 2 travels an extra distance with the vehicle to pick up rider 3, but that's only a very short distance compared to the original total trip distance, and it comes in exchange for a much cheaper ride. In the simulated scenario, we have already served 5 out of 6 requests with only two vehicles and thus our algorithm provides better fleet utilization.}
\section{Matching and Route-Planning}\label{apd:match}
\subsection{Algorithm for Potential Assignments}
The algorithm for deciding potential assignments is given in Algorithm \ref{assign}. 
\begin{algorithm}
	\caption{Potential Assignments} \label {assign}
	\fontsmall
	\begin{algorithmic}[ht]
		\State \textbf{Input: } Available Vehicles $V_{t}$ with their locations $loc(V_{j})$ such that: $V_{j} \in V_{t}$, Ride Requests $D_{t}$ with origin $o_{i}$ and destination $d_{i}$ associated with $r_{i} \in D_{t}$.
		\State \textbf{Output: } Matching decisions $A_{j}$ for each $V_{j} \in V_{t}$
		\State \textbf{Initialize} $A_{j} = [\;]$ for each $V_{j} \in V_{t}$.
		\For {each $r_{i} \in D_{t} \dots$}
		\State \textbf{Obtain} locations of candidate vehicles $V_{\text{cand}}$, such that:\\ $|loc(V_{j}) - o_{i}| \leq 5 \; km^{2}$.
		\State \textbf{Calculate} trip time $T_{j,i} \in T_{\text{cand}, i}$ from each $loc(V_{j}) \in V_{\text{cand}}$ to $o_{i}$ using the ETA model.
		\State \textbf{Pick} $V_{j}$ whose $T_{j,i} = \text{argmin}(T_{\text{cand}, i})$ to serve ride $r_{i}$.
		\State \textbf{Push} $r_{i}$ to $A_{j}$
		\EndFor
		\State \textbf{Return} $A_{t} = [A_{j}, A_{j+1}, ..., A_{n}]$, where $n = |V_{t}|$.
	\end{algorithmic} 
\end{algorithm}

\subsection{Complexity Analysis:} Here we discuss checking the route feasibility in $O(1)$. For a route to be feasible, for each request $r_{i}$ in this route, $o_{i}$ has to come before $d_{i}$. Therefore, to further reduce the computation needed, we first find the optimal position $Pos[o_{i}]$ to insert $o_{i}$ and then, find the optimal position $Pos[d_{i}]$ to insert $d_{i}$ we only consider positions starting from $Pos[o_{i}] + 1$. Therefore, we never have to check all permutations of positions, we only check $n^{2}$ options in the ride-sharing environment as we check the route feasibility in $O(1)$ time. This is further reflected in the capacity constraint defined in phase 1, where we borrow the idea of defining the smallest position to insert origin without violating the capacity constraint from \cite{6} as $Ps[l]$, the capacity constraint of vehicle $V_{j}$ will be satisfied if and only if: $Ps[l] \leq Pos[o_{i}]$.
Here, we need to guarantee that there does not exist any position $l \in (Pos[o_{i}], Pos[d_{i}])$ such that: $V^{j}_{C}[l] \; \geq \; \; (\mid r_{i-1} \mid - \mid r_{i} \mid)$, other than $Ps[l]$ to satisfy $Ps[l] \leq Pos[o_{i}]$ and thus abide by the capacity constraint.

\section{Reward Objectives for DQN}\label{apdx:reward}

In this section, we will define the different objectives for the reward function of DQN. 

\begin{enumerate}[leftmargin=*]
	\item Minimize the supply-demand mismatch, recall that $v_{t,m}$, and $\bar{d}_{t,m}$ denotes the number of available vehicles, and the anticipated demand respectively at time step $t$ in zone $m$. We want to minimize their difference over all $M$ zones, therefore, we get:
	\vspace{-.1in}
	\begin{equation}
		\fontsmall
		\text{diff}_{t} = \sum^{M}_{m = 1} (\bar{d}_{t,m} - v_{t,m})
	\end{equation}
	The reward will be learnt from the environment for individual vehicles, therefore, we map this term for individual vehicles. When vehicle serves more requests, the difference between supply and demand is minimized, and helps satisfy the demand of the zone it is located in. Therefore, we can get the total number of customers served by vehicle $n$ at time step $t$:
	\vspace{-.1in}
	\begin{equation}
		\begin{multlined}
			\fontsmall
			\text{C}_{t,n} = \sum^{M}_{m = 1} v^{n}_{t,m} \: (\text{where }v_{t,m} = 1 \text{ when } v_{t,m} < \bar{d}_{t,m}) \\
			\fontsmall \text{where } \sum^{M}_{m = 1} v^{n}_{t,m} = 1 \: \: (\gamma_{n,t,m} \in \{0,1\} \; \text{where } n \in v_{t,m})
		\end{multlined}
	\end{equation}
	\item Minimize the dispatch time, which refers to the expected travel time of vehicle $V_{j}$ to go zone $m$ at time step $t$, denoted by $h^{j}_{t,m}$. This can be estimated using ETA model. 
	%
	For individual vehicles, considering the neighboring vehicles' locations while making their decision, we get the total dispatch time, $T^{D}$ for vehicle $n$ at time step $t$:
	\vspace{-.1in}
	\begin{equation}
		\fontsmall
		T^{D}_{t, n} = \sum^{M}_{m = 1} h^{n}_{t,m} \; \; \; \; \{\text{where } \: n \in v_{t,m}\}
	\end{equation}
	\item Minimize the difference in times that the vehicle would have taken if it only serves one customer and the time it would take for car-pooling. For vehicles that participate in ride-sharing, an extra travel time may be incurred due to (1) either taking a detour to pickup an extra customer or (2) after picking up a new customer, the new optimal route based on all destinations might incur extra travel time to accommodate the new customers. This will also imply that customers already on-board will encounter extra delay. Therefore, that difference in time needs to be minimized, otherwise both customers and drivers would be disincentivized to car-pool. Let $t'$ be the total time elapsed after the passenger $l$ has requested the ride, $t_{n,l}$ be the travel time that vehicle $n$ would have been taken if it only served rider $l$, and $\tilde{t}_{n,l}$ be the updated time the vehicle $n$ will now take to drop off passenger $l$ because of the detour and/or picking up a new customer at time $t$. Note that $\tilde{t}_{n,l}$ is updated every time a new customer is added. Therefore, for vehicle $n$, rider $l$ at time step $t$, we want to minimize: $ \xi_{t,n,l} =  t' + \tilde{t}_{n,l} - t_{n,l}$. But for vehicle $n$, we want to minimize over all of its passengers, thus: $\sum^{\cup_{n}}_{l = 1} \xi_{t,n,l}$, where $\cup_{n}$ is the total number of chosen users for pooling at vehicle $n$ till time $t$. Note that $\cup_{n}$ is not known apriori, but will be adapted dynamically in the DQN policy. It will also vary as the passengers are picked or dropped by vehicle $n$. We want to optimize over all $N$ vehicles, therefore, the total extra travel time can be represented as:
	\vspace{-.1in}
	\begin{equation}
		\fontsmall
		\Delta t = \sum^{N}_{n = 1} \sum^{\cup_{n}}_{l = 1} \xi_{t,n,l}.
	\end{equation}
	For individual vehicles, the extra travel time for vehicle $n$ at time step $t$ becomes:
	\vspace{-.1in}
	\begin{equation}
		\fontsmall
		T^{E}_{t,n} = \sum^{\cup_{n}}_{l = 1} \xi_{t,n,l}.
	\end{equation}
	
	\item Maximize the fleet profits. This is calculated as the average earnings $E_{t}$ minus the average cost of all vehicles. Cost is calculated by dividing the total travel distance of vehicle $V_{j}$ by its mileage, and multiplied by the average gas price $P_{G}$. 
	Therefore, for individual vehicles, average profits for vehicle $n$ at time step $t$ is :
	\vspace{-.1in}
	\begin{equation}
		\fontsmall
		\mathbb{P}_{t,n} =  E_{t,n} - \left[ \frac{D_{t,n}}{M^{n}_{V}} * P_{G} \right]
	\end{equation}
	\item Minimize the number of utilized vehicles/resources. We capture this by minimizing the number of vehicles that become active from being inactive at time step $t$. Although we are minimizing the number of active vehicles in time step $t$, if the total distance or the total trip time of the passengers increase, it would be beneficial to use an unoccupied vehicle instead of having existing passengers encounter a large undesired delay. 
	Let $e_{t,n}$ represent whether vehicle $n$ is non-empty at time step $t$. The total number of vehicles that recently became active at time $t$ is given by:
	\vspace{-.1in}
	\begin{equation}
		\fontsmall
		e_{t} = \sum^{N}_{n = 1} \left[\text{max}(e_{t,n} - e_{t-1,n}, 0) \right]
	\end{equation}
\end{enumerate}
%
%

%
\section{DQN Architecture} \label{DQN_Arch}

\begin{figure}
	\vspace{-2.5em}
	\begin{center}
		\includegraphics[scale=0.3]{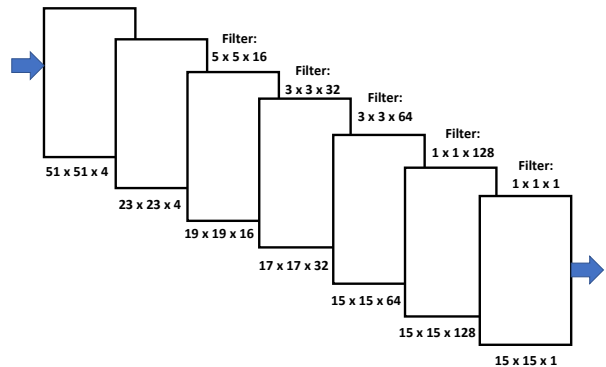}
		\vspace{-3em}
		\caption{The architecture of the Q-network. The output represents the Q-value for each possible movement/dispatch} \label{dqn_arch}
	\end{center}
\end{figure}

Figure \ref{dqn_arch} demonstrates the architecture of the Q-network. The output represents the Q-value for each possible movement/dispatch. In our simulator, the service area is divided into 43x44, cells each of size 800mx800m. The vehicle can move (vertically or horizontally) at most 7 cells, and hence the action space is limited to these cells. A vehicle can move to any of the 14 vertical (7 up and 7 down) and 14 horizontal (7 left and 7 right). This results in a 15x15 map for each vehicle as a vehicle can move to any of the 14 cells or it can remain in its own cell. The input to the neural network consists of the state representation, demand and supply, while the output is the Q-values for each possible action/move (15 moves). The input consists of a stack of four feature planes of demand and supply heat map images each of size 51x51. In particular, first plane includes the predicted number of ride requests next 30 minutes in each region, while the three other planes provide the expected number of available vehicles in each region in 0; 15 and 30 minutes. Before passing demand and supply images into the network, different sizes of average pooling with stride (1, 1) to the heat maps are applied, resulting in 23 x 23 x 4 feature maps. The first hidden layer convolves 16 filters of size 5x5 followed by a rectifier non-linearity activation. The second and third hidden layers convolve 32 and 64 filters of size 3x3 applied a rectifier non-linearity. Then, the output of the third layer is passed to another convolutional layer of size 15 x 15 x 128. The output layer is of size 15 x 15 x 1 and convolves one filter of size 1x1. Since reinforcement learning is unstable for nonlinear approximations such as the neural network, due to correlations between the action-value, we use experience replay to overcome this issue. Since every vehicle runs its own DQN policy, the environment during training changes over time from the perspective of individual vehicles.

\section{Learning Expected Discounted Reward Function, or  Q-values } \label{learn}
In our algorithm, we use reinforcement learning to learn the reward function stated in (\ref{individual}) using DQN. Through learning the probabilistic dependence between the action and the reward function, we learn the Q-values associated with the probabilities $P(r_{t}\mid a_{t},s_{t})$ over time by feeding the current states of the system. Instead of assuming any specific structure, our model-free approach learns the Q-values dynamically using convolutional neural networks whose architecture is described in Section \ref{arch}. Deep queue networks are utilized to dynamically generate optimized values. This technique of learning is characterized by its high adaptability to dynamic features in the system, which is why it is widely adopted in modern decision-making tasks. The optimal action-value function for vehicle $n$ is defined as the maximum expected achievable reward. Thus, for any policy $\pi_{t}$, we have:
\begin{equation}
\begin{multlined}
Q^{*}(s,a) = max_{\pi}\; \mathbb{E} \; \left[\;\sum^{\infty}_{k=t} \eta^{k-t} r_{k,n} \right.   \mid (s_{t,n} = s, a_{t,n} = a, \pi_{t}) \left. \right]
\end{multlined}
\end{equation}
where $0 < \eta < 1$ is the discount factor for the future. If $\eta$ is small (large, resp.), the dispatcher is more likely to maximize the immediate (future, resp.) reward. At any time slot $t$, the dispatcher monitors the current state $s_{t}$ and then feeds it to the neural network (NN) to generate an action. In our algorithm, we utilize a neural network to approximate the Q function in order to find the expectation.

For each vehicle $n$, an action is taken such that the output of the neural network is maximized. The learning starts with no knowledge and actions are chosen using a greedy scheme by following the Epsilon-Greedy method. Under this policy, the
agent chooses the action that results in the highest Q-value with probability $1-\epsilon$, otherwise, it selects a random action. The $\epsilon$ reduces linearly from 1 to 0.1 over $T_{n}$ steps. For the $n^{th}$ vehicle, after choosing the action and according to the reward $r_{t,n}$, the Q-value is updated with a learning factor $\sigma$ as follows:
\begin{equation} \label{Qupdate}
\begin{multlined}
Q^{'}(s_{t,n},a_{t,n})\; \leftarrow \; (1 - \sigma) \; Q(s_{t,n},a_{t,n}) \; \\ + \;
\sigma \; [r_{t,n} + \eta \; \text{max}_{a} \; Q(s_{t+1,n},a)\; ]
\end{multlined}
\end{equation}
Similar to $\epsilon$, the learning rate $\sigma$ is also reduced linearly from 0.1 to 0.001 over 10000 steps. We note that an artificial neural network is needed to maintain a large system space. When updating these values, a loss function $L_{i}(\theta_{i})$ is used to compute the difference between the predicted Q-values and
the target Q-values, i.e.,
\begin{equation}
\begin{multlined}
L_{i}(\theta_{i}) \; = \; \mathbb{E}  \; \left[\; \left( (r_{t} \; + \; \eta \; \text{max}_{a} Q(s,a; \bar{\theta_{i}})) \right. \right. - Q(s,a; \theta_{i}) \left. \right)^{2} \; \left. \right]
\end{multlined}
\end{equation}
where $\theta_{i}, \bar{\theta_{i}}$, are the weights of the neural networks. This above expression represents the mean-squared error in the Bellman equation where the optimal values are approximated with a target value of $r_{t} \; + \; \eta \; \text{max}_{a} Q(s,a; \bar{\theta_{i}})$, using the weight $\bar{\theta_{i}}$ from some previous iterations.

\end{document}